\documentclass[pdflatex,Numbered,sn-basic]{cheminfo/sn-jnl}
\usepackage{graphicx}%
\usepackage{multirow}%
\usepackage{amsmath,amssymb,amsfonts}%
\usepackage{amsthm}%
\usepackage{xcolor}%
\usepackage{textcomp}%
\usepackage{manyfoot}%
\usepackage{booktabs}%
\usepackage{array}%
\usepackage{longtable}%
\usepackage{algorithm}%
\usepackage{algorithmicx}%
\usepackage{algpseudocode}%
\usepackage{listings}%
\usepackage{adjustbox}
\usepackage{subcaption}
\usepackage{placeins}
\usepackage{float}
\usepackage{xstring}

\definecolor{darkred}{rgb}{0.9, 0, 0}

\newcommand{\prompt}[1]{%
    \begingroup
    \colorlet{numbers}{blue}%
    \colorlet{tags}{darkred}%
    \texttt{#1}%
    \endgroup
}
\newcommand{\rev}[1]{\textcolor{black}{#1}}

\theoremstyle{thmstyleone}%
\theoremstyle{thmstyletwo}%

\theoremstyle{thmstylethree}%

\begin{document}

\title[Small Molecule Optimization with LLMs]{Small Molecule Optimization with Large Language Models}


\author[1,2]{\fnm{Philipp} \sur{Guevorguian}}

\author[1,2]{\fnm{Menua} \sur{Bedrosian}}

\author[1,2]{\fnm{Tigran} \sur{Fahradyan}}

\author[1]{\fnm{Gayane} \sur{Chilingaryan}}

\author{\fnm{Armen} \sur{Aghajanyan}}

\author*[1,2]{\fnm{Hrant} \sur{Khachatrian}}\email{hrant@yerevann.com}

\affil[1]{\orgname{YerevaNN research lab},  \city{Yerevan}, \country{Armenia}}

\affil[2]{\orgname{Yerevan State University}, \city{Yerevan}, \country{Armenia}}


\abstract{Molecular optimization, the process of designing molecules with desirable properties, represents a critical challenge in drug discovery.
The recent advancements in large language models (LLMs) have opened new opportunities for their integration with traditional molecular optimization algorithms to improve performance. In this work, we propose \textbf{Mo}lecular \textbf{L}anguage Model powered \textbf{E}volutionary Algorithm (Mol-E), an evolutionary algorithm that relies on the generative capabilities of LLMs trained on molecules and molecular properties.

\textbf{Scientific Contribution.} \rev{Mol-E obtains the highest aggregate Top-10 AUC among the comparable full-23-task results considered here, scoring 17.500 in the task-agnostic regime, in which the oracle is treated strictly as a black box, and 20.551 in the task-informed regime, in which the optimizer receives a fixed semantic description of the objective.} Mol-E also improves over the evaluated baselines on multi-property optimization with docking against DRD2, MK2, and AChE.

}




\maketitle

\section{Introduction}\label{sec:intro}

Molecular optimization is a fundamental challenge in drug discovery. It requires searching in a combinatorially large chemical space, with a widely cited estimate placing the number of possible small organic molecules above $10^{60}$ \citep{bohacek1996art}, and identifying molecules with desired therapeutic properties. This process traditionally required extensive laboratory experiments, making it time-consuming and costly. Computational methods have emerged as powerful tools to accelerate this process. Particularly, deep learning techniques have been used to accelerate the process of molecular design and predicting molecular properties \citep{survey-drug-discovery-prog-chal-opps, survey-raise-dl-drug-discovery, survey-small-mol-drug-discovery}. \\

\begin{figure}[H]
    \centering
    \includegraphics[width=1\linewidth]{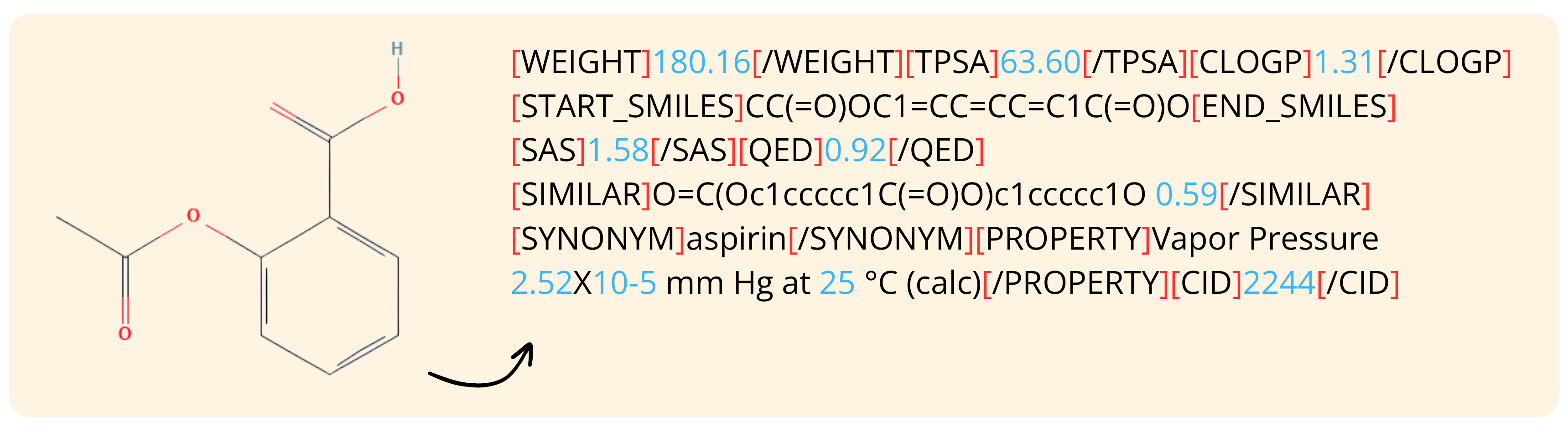}
    \caption{Aspirin's molecular structure (left) and its representation in our dataset (right)}
    \label{fig:aspirin}
\end{figure}

\noindent LLMs have demonstrated remarkable results in various domains and have opened new opportunities to use them in designing new molecular optimization algorithms \citep{gpt4, gemini, BERT, BART}. While LLMs are powerful generative models, integrating them in powerful molecular optimization algorithms remains a key area of research. \\

\noindent A particular class of approaches that stand out are LLM-enhanced evolutionary algorithms, which have demonstrated breakthroughs in discovering new mathematics and algorithm design \citep{alphaevolve, funsearch}. To address the open question of designing optimization algorithms with LLMs and motivated by the success of combining evolutionary algorithms and LLMs, we propose a novel evolutionary search framework that uses the generative power of LLMs trained on molecular corpora, which we call \textbf{Mo}lecular \textbf{L}anguage Model powered \textbf{E}volutionary Algorithm (Mol-E). Specifically, Mol-E utilizes the understanding and generative power of the LLMs about molecules and molecular properties to effectively solve molecular optimization tasks. \\

\noindent We evaluate Mol-E on the challenging Practical Molecular Optimization (PMO) benchmark \citep{pmo} under two explicitly separated information regimes. In the task-agnostic regime, the optimizer treats each oracle as a black box and the best Mol-E variant reaches a summed Top-10 AUC of 17.500 across the 23 tasks. In the task-informed regime, fixed prompts describe objectives using molecular-property tags learned during pretraining, and the best variant reaches 20.551. We additionally demonstrate Mol-E on multi-property optimization problems involving protein--ligand docking, showcasing its relevance to realistic drug-discovery settings.

\section{Related Work}
\subsection{Language Models for Molecular Representation}
String-based representations, particularly Simplified Molecular Input Line Entry System (SMILES) \citep{smiles} and Self-Referencing Embedded Strings (SELFIES) \citep{selfies}, have gained new traction in molecular modeling due to their compatibility with language modeling approaches. Standard language modeling techniques have been applied for molecular representation learning, such as sequence-to-sequence models \citep{smiles-rnn, smiles-transformer, chemformer, molgpt, bartsmiles, molt5} and transformer encoder-only models \citep{smiles-bert, molformer-nature, chemberta2, selformer}. These models demonstrate the potential of applying natural language processing techniques to molecular design and property prediction.

\subsection{Molecular Optimization Algorithms}
Molecular optimization involves navigating a vast combinatorial space of potential drugs while satisfying multiple constraints. Traditional molecular optimization methods include adaptation of evolutionary algorithms and Monte Carlo tree search to molecular graphs \citep{evo-alg-opt-survey, smiles-ga, graph-GA}. Recent methods increasingly make use of machine learning, particularly deep learning techniques \citep{van2024deep}. Variational autoencoders \citep{vae} have been applied to generate and optimize molecules \citep{smiles-vae, jin2018junction}. Reinforcement learning (RL) techniques incorporating language models have been extensively developed for molecular design \citep{transformer-policy-grad, denovo-policy-grad}, notably REINVENT \citep{blaschke2020reinvent}, and its extensions utilizing diversity filters, experience replay mechanism \citep{augmented-memory, beam-enumeration}, and genetic search \citep{saturn}. The Generative Flow Networks (GFlowNets) \citep{gflownet} represent a relatively recent and powerful approach for sampling molecules, which demonstrated promising results in molecular optimization \cite{goal-cond-gflownet, genetic-gflownet, atomic-gflownet}.

\rev{Recent PMO studies span black-box search with learned surrogates or generative models, including LICO \citep{lico}, GPBO \citep{gpbo}, ACEGEN \citep{acegen}, GP-MoLFormer-Sim+GA \citep{gp-molformer}, and InVirtuoGen \citep{invirtuogen}, and language-model agents that receive semantic task information, including MOLLEO \citep{molleo}, ExLLM \citep{exllm}, MT-Mol \citep{mt-mol}, MolLIBRA \citep{mollibra}, FORGE \citep{forge}, INDIBATOR \citep{indibator}, and SEISMO \citep{seismo}. Other work studies oracle-specific online training or synthesis-aware search, including NovoMolGen \citep{novomolgen}, ChemTSv3 \citep{chemtsv3}, SynGBO \citep{syngbo}, S3-GFN \citep{s3gfn}, short-term graph memory \citep{stgm}, and SEGO \citep{sego}. Because these studies differ in oracle-call horizon, task coverage, semantic information, oracle implementation, and whether target-oracle prescreening is charged to the budget, we report their results with explicit protocol qualifications.}

\rev{Fragment-based f-RAG \citep{frag} and GenMol \citep{genmol} construct oracle-specific fragment vocabularies by scoring ZINC250k before their reported trajectories. Graph-GRPO \citep{graph-grpo} trains a separate graph-flow policy against each PMO oracle before evaluation, and the prescreened InVirtuoGen variant \citep{invirtuogen} likewise uses target-oracle-scored molecules before its measured run. We therefore distinguish these methods from cold-start comparisons because their target-oracle prescreening or policy-training calls are omitted from the nominal budget.}

\subsection{Evolutionary Algorithms in Molecular Optimization}

The synthesis of evolutionary algorithms with deep learning models gave rise to powerful methods for solving challenging combinatorial optimization problems and has demonstrated remarkable results in various critical problems \citep{funsearch, alphaevolve, funsearch-graph-sim-dist, llms-as-optimizers, evoprompting, evoprompt}. Our work builds on top of these successes, proposing to design molecules using evolutionary algorithms augmented by the generative capabilities of LLMs. The closest existing works to ours are \citep{gegl}, which combines language models with genetic search to design molecules and \citep{rnn-genetic, molleo}, which use language models to implement the crossover and mutation operations of the genetic algorithm for optimizing molecules. SEISMO \citep{seismo} instead uses Claude Opus 4.5 as a trajectory-conditioned agent supplied with natural-language task descriptions and explanatory oracle feedback.

\section{Methods}
In this section, we discuss methodology and design choices for training our language models fluent in the language of molecular properties, the construction of our evolutionary algorithm, and other capabilities that are achievable given our approach.

\subsection{Training a Language Model for Molecules}
Here we present our family of chemistry-specialized models Chemlactica (125M and 1.3B parameters) and Chemma (2B parameters), based on Galactica \citep{taylor2022galactica} and Gemma \citep{team2024gemma} respectively, trained on a comprehensive molecular corpus derived from PubChem \citep{pubchem}, which includes 110 million molecules and their properties. We chose models for continued pretraining based on their general-purpose performance and domain-specific knowledge. At its release, Galactica outperformed models like OPT \citep{opt}, and Chinchilla \citep{chinchilla} on tasks such as BIG-bench \citep{bigbench}, and MMLU \citep{mmlu}. Its pretraining included two million PubChem molecules, SMILES-specific tagging, and a scientific corpus, making it well-suited for molecular data. Gemma, while not explicitly trained on molecular data, underwent extensive pretraining (2 trillion tokens for Gemma-2B) and demonstrated state-of-the-art performance on benchmarks like MMLU, and HellaSwag \citep{hellaswag}.
We used the Galactica and Gemma tokenizers with minor modifications, and continued pretraining with a causal language modeling objective. Additional details on tokenization and training are provided in Supplementary Section \ref{sec:training_details}.

\subsubsection{Continued Pretraining}
\rev{To assess the advantage of initializing from a pretrained model, we randomly initialized a 125M-parameter Galactica model and trained it on our complete corpus following the same protocol. Table \ref{tab:cg-pp} shows that this randomly initialized model performs worse than Chemlactica-125M on every evaluated property-prediction and conditional-generation task.}

\subsubsection{Training Corpus}
\label{sec:data_main}
For our training data, we extract information on molecules from PubChem \citep{pubchem}, encompassing information on the molecules and molecular properties. We reserve around 10,000 molecules as a validation set to monitor during the training. We used RDKit \citep{RDKIT} to compute key molecular properties, compute more precise similarity measurements, and standardize SMILES strings. To provide this as input to the language models, we developed a language using paired tags to delimit each property and data point for the final string representation of molecules. For example, the string representation for the quantitative estimate of drug-likeness (QED) value of a molecule is \prompt{[QED]0.84[/QED]}. To enable property prediction and property-conditioned molecular generation, we randomized the property order and either set the position of the primary molecule at the start of the document or between other tags with equal probability. Figure \ref{fig:aspirin} illustrates the document corresponding to aspirin. We provide more details on the training data in Supplementary Section \ref{sec:data_added}.

\subsubsection{Language Model Capabilities}
Given the inherent flexibility of language models that comes with prompting, our chemistry-aware models can be employed in different setups. In the following sections, we describe how our models serve as the generative component of our evolutionary algorithm, as well as a property predictor for molecules. Furthermore, we show how the model is able to generate new molecules conditioned on different properties, as well as adapt to new datasets and downstream tasks through supervised fine-tuning.

\subsection{Language Model Powered Evolutionary Algorithm}

We introduce a novel population-based algorithm that effectively navigates the vast chemical space to identify molecules with targeted properties, which we call \textbf{Mo}lecular \textbf{L}anguage Model Powered \textbf{E}volutionary Algorithm (Mol-E). Mol-E incorporates the generative power of LLMs into the evolutionary search framework.

\subsubsection{Evolutionary Algorithm and Prompting}

Mol-E maintains a dynamic pool of $P$ high-performing molecules. In each iteration, this pool serves as the parent population for generating new offspring candidates. We leverage the language model's similarity-conditioned molecule generation capability for this process, which can be interpreted to implement the crucial operations of an evolutionary algorithm, such as crossover and mutation. \\

\noindent Mol-E starts off by initializing an empty pool of molecules. At first iteration of the evolutionary algorithm the pool is populated with random molecules from the model and at each subsequent iteration $S$ random molecules are selected from the pool to serve as an inspiration for the offspring and placed in the prompt with similarity conditions as follows \\

\noindent \small{\prompt{[SIMILAR]}$m^{smiles}_1$\prompt{ $v^{sim}_{1}$}\prompt{[/SIMILAR]...[SIMILAR]}$m^{smiles}_S$\prompt{ $v^{sim}_{S}$[/SIMILAR][START\_SMILES].}} \\

\noindent The model generates new molecules from this prompt, which are then scored against the oracle. The newly generated molecules are merged with the pool, and duplicates are removed. Each evolutionary iteration concludes by retaining the top-$P$ highest-scoring molecules and discarding the remainder, after which the next iteration begins with the updated pool. The optimization terminates when the oracle budget is reached.



\subsubsection{Dynamic Fine-Tuning During the Optimization}

\noindent To directly incorporate the oracle feedback and prevent the optimization from stalling, we employ a dynamic fine-tuning strategy. When the progress of the algorithm stagnates, we fine-tune the language model using the highest-scoring molecules and their corresponding oracle scores. This explicit modeling allows the language model to learn the relationship between molecular structures and oracle scores, enabling more targeted generation in subsequent iterations. The fine-tuning uses samples of the form: 
\noindent\small{\prompt{[PROPERTY]$O(m)$[/PROPERTY][SIMILAR]$m^{smiles}_1$\prompt{ $v^{sim}_{1}$}\prompt{[/SIMILAR]...[SIMILAR]}$m^{smiles}_S$\prompt{ $v^{sim}_{S}$[/SIMILAR]}
\prompt{[START\_SMILES]$m^{smiles}$[END\_SMILES]}}}
\noindent where $m^{smiles}$ is a generated molecule drawn from the pool, $O(m)$ is its previously computed oracle score, and the remaining molecules provide similarity conditions. Fine-tuning is triggered when the highest score in the pool does not improve for $K$ consecutive evolutionary iterations. The examples are constructed from all molecules in the pool using their existing oracle scores, so fine-tuning consumes no additional oracle evaluations. After the first fine-tuning step, generation prompts incorporate the learned oracle signal by appending \small{\prompt{[PROPERTY]$v^{oracle}$[/PROPERTY]}}.

\noindent We present the complete optimization procedure in Algorithms \ref{alg:molecular_optimization}, \ref{alg:molecules2prompt}, and \ref{alg:molecules2sample}, which implement the main optimization loop, the prompt construction, and fine-tuning sample construction processes, respectively. \\


\noindent Given the complexity of our algorithm, we adopt a focused hyperparameter tuning strategy, prioritizing the most sensitive parameters while keeping others fixed. This approach balances computational efficiency with optimization performance. Supplementary Section \ref{subsec:pmo-hparams} provides the methodology and results of our hyperparameter tuning experiments.

\subsection{Chemistry-Specialized Language Model Capabilities}

\subsubsection{Property Prediction}
We randomly sampled a fixed set of 100 molecules from the validation set. For each property, we prompted the models with \prompt{[START\_SMILES]$m^{smiles}_{i}$[END\_SMILES][QED]}, where $m^{smiles}_{i}$ represents the SMILES string of the molecule. We then calculated the root mean square error between predicted and actual property values to evaluate performance.
\subsubsection{Conditional Generation}
For each property, we sampled 100 values $v_i$ from the distribution of PubChem molecules. We then prompted the models to generate molecules with \prompt{[QED]$v_i$[/QED][START\_SMILES]}. Using RDKit, we computed the actual property values of the generated SMILES and calculated the RMSE against the target $v_i$. To account for potential invalid generations, we compute a corrected RMSE by substituting the property values of invalid SMILES with the mean value of the respective property's distribution in our dataset.
Table \ref{tab:cg-pp} demonstrates the results of these experiments for the three models and two controlled versions of the Chemma-2B model:  (Chemma-2B-2.1B), which retains the 2B-parameter architecture but is trained on 2.1B tokens for the compute-controlled experiment, and  (Chemma-2B-39B), which similarly retains the 2B-parameter architecture but is trained on 39B tokens to match the number of molecules seen by the 125M model trained on the full training set. We have utilized several techniques, including Chain-of-Thought (CoT) \citep{chainofthought}, repetition penalty \citep{ctrl}, and undesired token suppression to enhance the quality of generations. The details of these techniques, alongside an ablation study of the effect of each on generations, are included in Supplementary Section \ref{sec:CGdetails}. Furthermore, we show that our models are well calibrated in predicting these properties in \ref{subsec:calibration}. These experiments comprehensively demonstrate our models' capabilities in predicting molecular properties and generating molecules with specified properties. These are crucial tasks in molecular design and will become the building blocks for our optimization algorithm.
\subsubsection{Supervised Fine-Tuning}
\rev{We further evaluate whether the models can learn assay-derived properties through supervised fine-tuning on the six single-task ADME benchmarks introduced by Fang et al. \citep{admet-industrial}. We use the official Polaris training and held-out test splits \citep{cas_wognum_2024_13652588} for human and rat liver microsomal stability, MDR1--MDCK efflux ratio, solubility, and human and rat plasma protein binding. Each model predicts the property value autoregressively through next-token prediction, without a dedicated regression head. Results are reported in Table \ref{tab:adme}; the fine-tuning recipe and selected hyperparameters are provided in Supplementary Section \ref{sec:property-predicion-apdx}.}

\begin{algorithm}[tb]
   \caption{ }
   \label{alg:molecular_optimization}
\begin{algorithmic}
    \State {\bfseries Input:} $P$, $S$, $N$, $K$
    \State 1. Initialize an empty
    \State $Pool \leftarrow \{\}$
    \Repeat
    \State 2. Generate prompts for molecule generation
    \For{$i=1$ {\bfseries to} $N$}
    \State $(m_{i, 1}, m_{i, 2}, \ldots, m_{i, S}) \leftarrow random\_subset(Pool)$
    \State $p_i \leftarrow \textbf{molecules2prompt}(m_{i, 1}, m_{i, 2}, \ldots, m_{i, S})$
    \EndFor
    \State
    \State 3. Generate $N$ new and unique molecules with the language model
    \State $m_{i} \leftarrow LM(p_{i}), i=1, \ldots, N$
    \State
    \State 4. Update the pool with the new candidates and keep only the top-$P$ molecules
    \State $Pool \leftarrow Pool \cup \{m_{1}, \ldots, m_{N}\}$
    \State $Pool \leftarrow $ top-$P(Pool)$
    \State
    \State 5. Fine-tune if necessary
    \If{the best oracle score has not improved for $K$ consecutive iterations}
    \State 5.1. Construct training samples for all the molecules in the $Pool$ with their corresponding similar molecules (using which they have been generated)
    \State $samples_{i} \leftarrow \textbf{molecules2sample}((m_{i, 1}, m_{i, 2}, \ldots, m_{i, S}), m_i), i=1, \ldots, P$
    \State
    \State 5.2. Train $LM$ on $samples_{i}, i=1, \ldots, P$.
    \EndIf
    \Until{optimization problem stopping condition}
\end{algorithmic}
\end{algorithm}

\medskip
\noindent A large language model was used to assist with language editing and manuscript formatting. The authors reviewed the resulting text and take full responsibility for the manuscript's content.
\medskip

\section{Results and Discussion}\label{sec:results}

To evaluate the molecular design capabilities of the Mol-E framework, we test it on the Practical Molecular Optimization (PMO) \citep{pmo} benchmark and on multi-property optimization tasks with molecular docking.

\subsection{Practical Molecular Optimization}

Inspired by real-world molecular design settings \cite{pmo} proposes the Practical Molecular Optimization (PMO) benchmark consisting of 23 molecular optimization problems. PMO focuses on sample efficiency, generalizability to different optimization objectives, and robustness to hyperparameter
selection of molecular optimization algorithms. To assess optimization power and sample efficiency, PMO evaluates the area under the curve (AUC) of the top-$10$ average property values over a 10,000-call horizon. AUC values are calculated after every 100 oracle calls, combined, and normalized to the $[0, 1]$ range. A run may terminate early under the benchmark's \texttt{finish=True} convention, in which case its terminal Top-10 value is held constant over the remainder of the evaluation horizon. \\

PMO results can be separated into two information regimes. In the \emph{task-agnostic} regime, the oracle is treated as a black box: an optimizer observes only molecules and their scalar scores, without receiving the oracle name, its chemical interpretation, or its component objectives. All methods evaluated in the original PMO study follow this regime \citep{pmo}. More recent work has also studied a \emph{task-informed} regime, in which the optimizer receives information about the objective beyond scalar oracle evaluations. This is particularly natural for language-model-based optimizers: MOLLEO supplies objective-specific natural-language descriptions to its editing models \citep{molleo}, and SEISMO conditions an LLM agent on task descriptions and the optimization trajectory \citep{seismo}. To our knowledge, MOLLEO was the first PMO study to explicitly expose such semantic task descriptions to the optimizer. We evaluate Mol-E in both regimes and report them separately. \\

For task-informed Mol-E, we prepend each generation prompt with a fixed task description expressed only through property tags seen during model pretraining. For example, the \texttt{ranolazine\_mpo} prefix specifies the target molecule and desired TPSA and CLogP values as \nolinkurl{[SIMILAR]COc1ccccc1OCC(O)CN1CCN(CC(=O)Nc2c(C)cccc2C)CC1\ 0.70[/SIMILAR][TPSA]95.00[/TPSA][CLOGP]7.00[/CLOGP]}. We do not select prompts using oracle evaluations and use the same registry and optimization hyperparameters across all 23 tasks for a given model. See Supplementary Section \ref{sec:pmo-task-informed-prompts} for details. \\

\rev{Table \ref{tab:PMO} compares the task-agnostic Mol-E reproductions with the original PMO baselines and subsequent PMO-10K methods. It also shows the three highest-scoring methods that remain task-agnostic with respect to semantic objective information but use uncounted target-oracle evaluations; these are displayed in gray and are not included in the comparable ranking. Table \ref{tab:PMO2} compares task-informed Mol-E with MOLLEO \citep{molleo}, ExLLM \citep{exllm}, and SEISMO \citep{seismo}. Supplementary Section \ref{sec:pmo-detailed-results} provides all task-level results and protocol qualifications, including the complete budget-noncomparable comparison in Table \ref{tab:PMO_full_budget_noncomparable}.} The hyperparameters are tuned for each Mol-E backbone separately according to Supplementary Section \ref{subsec:pmo-hparams}. For these experiments, we load the model parameters in full fp32 precision. \\

\begin{table}[h]
\begingroup
\centering
\resizebox{\columnwidth}{!}{
\begin{tabular}{@{}lcccccc@{}}
\toprule
                             & jnk3 &  median1  & scaffold\_hop & sitagliptin\_mpo & sum of 4 & sum of 23 \\ \midrule
\rev{SEISMO (no description)$^{\ddagger}$ \citep{seismo}} & \rev{0.137} & \rev{0.131} & \rev{0.459} & \rev{0.176} & \rev{0.903} & \rev{10.019} \\
\rev{SEGO$^{\ddagger}$ \citep{sego}} & \rev{0.055} & \rev{0.276} & \rev{0.508} & \rev{0.222} & \rev{1.061} & \rev{11.410} \\
REINVENT \citep{pmo} & 0.783 & 0.356 & 0.560 & 0.021 & 1.720 & 14.196 \\
\rev{S3-GFN \citep{s3gfn}} & \rev{0.797} & \rev{0.285} & \rev{0.561} & \rev{0.451} & \rev{2.094} & \rev{14.255} \\
\rev{LICO \citep{lico}} & \rev{0.731} & \rev{0.291} & \rev{0.547} & \rev{0.567} & \rev{2.136} & \rev{14.708} \\
Augmented Memory \citep{augmented-memory} & 0.739 & 0.326 & 0.567 & 0.284 & 1.916 & 15.002 \\
\rev{GP-MoLFormer-Sim+GA \citep{gp-molformer}} & \rev{0.806} & \rev{0.340} & \rev{0.531} & \rev{0.501} & \rev{2.178} & \rev{15.225} \\
\rev{GenMol+STGM$^{\S}$ \citep{stgm}} & \rev{0.840} & \rev{0.349} & \rev{0.559} & \rev{0.433} & \rev{2.181} & \rev{15.664} \\
\rev{ACEGEN-PPOD \citep{acegen}} & \rev{0.870} & \rev{0.350} & \rev{\textbf{0.840}} & \rev{0.390} & \rev{2.450} & \rev{15.800} \\
\rev{ChemTSv3$^{\dagger}$ \citep{chemtsv3}} & \rev{0.650} & \rev{0.340} & \rev{0.550} & \rev{\textbf{0.780}} & \rev{2.320} & \rev{16.100} \\
Genetic-guided GFlowNet \citep{genetic-gflownet} & 0.764 & 0.379 & 0.615 & 0.634 & 2.392 & 16.213 \\
\rev{GPBO \citep{gpbo}} & \rev{0.785} & \rev{\textbf{0.415}} & \rev{0.529} & \rev{0.474} & \rev{2.203} & \rev{16.303} \\
\rev{SynGBO \citep{syngbo}} & \rev{0.910} & \rev{0.357} & \rev{0.541} & \rev{0.454} & \rev{2.262} & \rev{16.425} \\
\rev{InVirtuoGen (cold start) \citep{invirtuogen}} & \rev{0.825} & \rev{0.342} & \rev{0.589} & \rev{0.709} & \rev{2.465} & \rev{16.676} \\
\rev{NovoMolGen \citep{novomolgen}} & \rev{0.818} & \rev{0.374} & \rev{0.790} & \rev{0.673} & \rev{\textbf{2.655}} & \rev{16.700} \\ \midrule
Mol-E (Chemlactica-125M) & 0.877 & 0.359 & 0.668 & 0.603 & 2.507 & 16.945 \\
Mol-E (Chemlactica-1.3B) & 0.854 & 0.361 & 0.617 & 0.539 & 2.372 & 17.169 \\
Mol-E (Chemma-2B) & \textbf{0.914} & \rev{0.377} & \rev{0.709} & 0.515 & \rev{2.515} & \textbf{17.500} \\ \midrule
\textcolor{gray}{Graph-GRPO, no prescreen$^{*}$ \citep{graph-grpo}} & \textcolor{gray}{0.910} & \textcolor{gray}{0.388} & \textcolor{gray}{0.622} & \textcolor{gray}{0.879} & \textcolor{gray}{2.799} & \textcolor{gray}{18.987} \\
\textcolor{gray}{InVirtuoGen, prescreened$^{*}$ \citep{invirtuogen}} & \textcolor{gray}{0.898} & \textcolor{gray}{0.386} & \textcolor{gray}{0.711} & \textcolor{gray}{0.743} & \textcolor{gray}{2.738} & \textcolor{gray}{18.993} \\
\textcolor{gray}{Graph-GRPO, prescreened$^{*}$ \citep{graph-grpo}} & \textcolor{gray}{0.910} & \textcolor{gray}{0.388} & \textcolor{gray}{0.711} & \textcolor{gray}{0.879} & \textcolor{gray}{2.888} & \textcolor{gray}{19.270} \\ \bottomrule
\end{tabular}}
\caption{Task-agnostic PMO comparison, sorted by the sum over all 23 tasks. Entries are mean Top-10 AUC $\uparrow$ values. Mol-E results average five independent seeds using the corrected normalized Sitagliptin oracle; baseline results follow their cited sources. \rev{$^{*}$Gray methods use no semantic task description but are not budget-comparable: Graph-GRPO omits target-oracle policy-training calls, and the prescreened Graph-GRPO and InVirtuoGen variants additionally omit approximately 250,000 target-oracle calls. $^{\dagger}$ChemTSv3 uses changed implementations of at least the isomer, Sitagliptin, and Zaleplon oracles. $^{\ddagger}$SEGO and SEISMO make 100 and 50 proposals, respectively, after which their terminal Top-10 values are held fixed through 10,000 calls. $^{\S}$GenMol+STGM reports one seed at 10,000 calls.}}
\label{tab:PMO}
\endgroup
\end{table}

\begin{table}[h]
\begingroup
\centering
\resizebox{\columnwidth}{!}{
\begin{tabular}{@{}lcccccc@{}}
\toprule
                             & jnk3 & median1 & scaffold\_hop & sitagliptin\_mpo & sum of 4 & sum of 23 \\ \midrule
MOLLEO (MolSTM) \citep{molleo} & 0.643 & 0.298 & 0.527 & 0.548 & 2.016 & 14.557 \\
MOLLEO (BioT5) \citep{molleo} & 0.728 & 0.338 & 0.559 & 0.506 & 2.131 & 15.424 \\
MOLLEO (GPT-4) \citep{molleo} & 0.790 & 0.352 & 0.971 & 0.584 & 2.697 & 17.862 \\
SEISMO (extrapolated) \citep{seismo} & 0.468 & 0.344 & 0.934 & 0.238 & 1.985 & 18.214 \\
\rev{ExLLM \citep{exllm}} & \rev{0.732} & \rev{0.384} & \rev{0.916} & \rev{0.555} & \rev{2.587} & \rev{19.165} \\ \midrule
Mol-E (Chemlactica-1.3B) & 0.854 & \textbf{0.489} & 0.986 & 0.704 & 3.033 & 20.192 \\
Mol-E (Chemlactica-125M) & 0.877 & 0.465 & \textbf{0.989} & 0.720 & 3.051 & 20.429 \\
Mol-E (Chemma-2B) & \textbf{0.912} & 0.482 & 0.983 & \textbf{0.739} & \textbf{3.116} & \textbf{20.551} \\
\bottomrule
\end{tabular}}
\caption{Task-informed PMO comparison. Entries are mean Top-10 AUC $\uparrow$ values. Mol-E, MOLLEO, and ExLLM average five independent seeds. \rev{SEISMO makes 50 proposals; we reconstruct its 10,000-call AUC by holding each released trajectory's terminal Top-10 value fixed through the remaining horizon.}}
\label{tab:PMO2}
\endgroup
\end{table}

For several QED seeds, Mol-E reached the maximum achievable Top-10 value early and subsequently failed to generate novel molecules. We stopped those runs early and flat-filled their terminal Top-10 values to a 10,000-call horizon when calculating AUC, following the benchmark's \texttt{finish=True} convention. \rev{For SEISMO, which makes 50 proposals and releases raw trajectories, we apply the same flat-fill rule to reconstruct AUC over the full horizon.}

\rev{The three highest-scoring budget-noncomparable methods are included in gray in Table \ref{tab:PMO}, while Supplementary Table \ref{tab:PMO_full_budget_noncomparable} gives the complete comparison. f-RAG, GenMol, and the prescreened InVirtuoGen variant use large-scale target-oracle prescreening of ZINC250k before the measured trajectory, while Graph-GRPO trains a separate policy against each target oracle before evaluation; these oracle evaluations are omitted from the nominal 10,000-call budget.}

The \texttt{valsartan\_smarts} oracle contains a binary multiplier that provides no optimization signal in most evaluations. Only a small number of task-agnostic trajectories escape this zero-reward region, whereas task-informed prompting is particularly helpful for this task.

\begin{figure}
    \centering
    \includegraphics[width=0.85\linewidth]{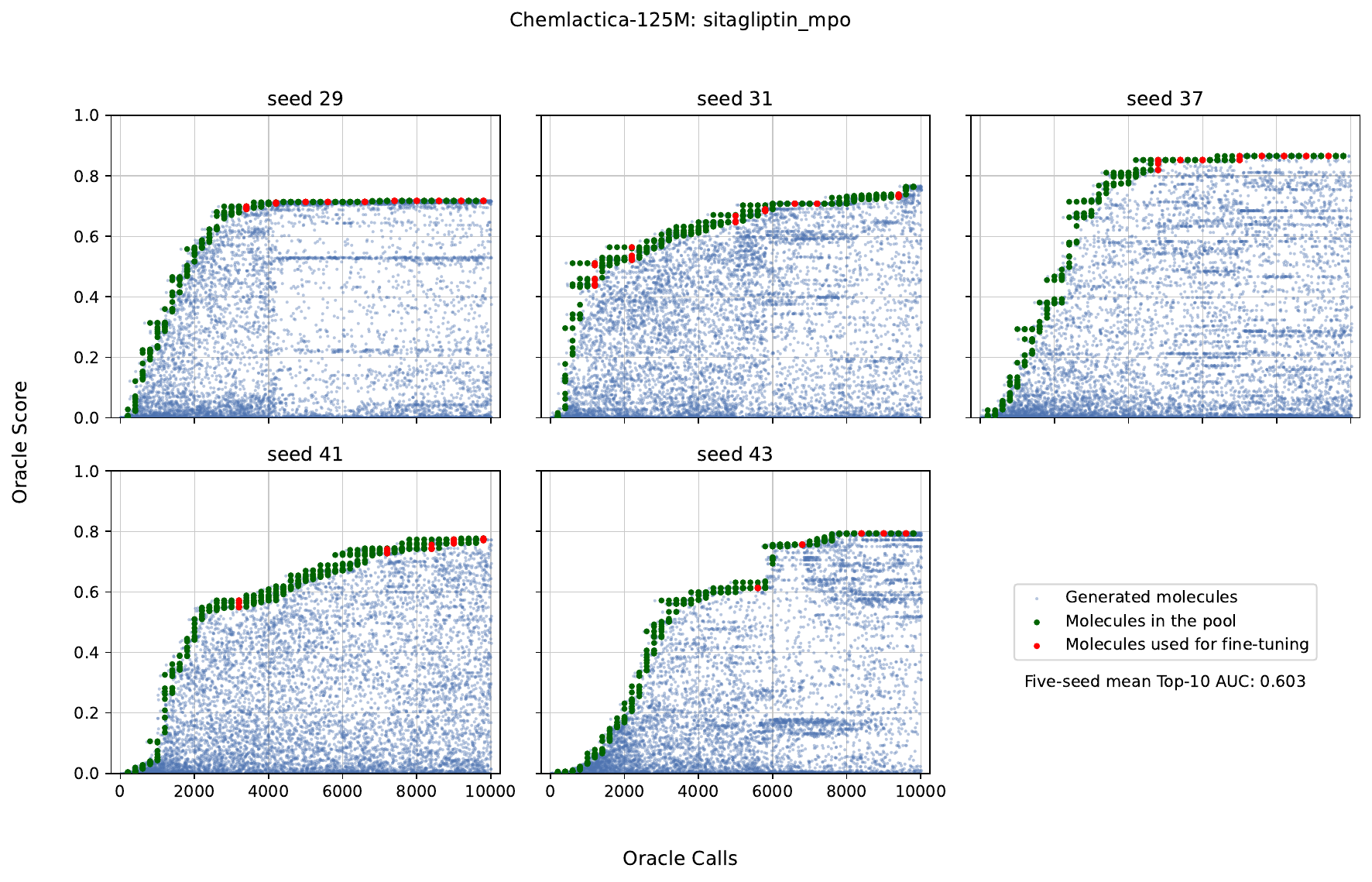}
    \caption{Optimization process visualization using Chemlactica-125M for the \texttt{sitagliptin\_mpo} task, using the five independent reproduction seeds.}
    \label{fig:Chemlactica-125M-sitagliptin_mpo}
\end{figure}

\begin{figure}
    \centering\includegraphics[width=0.85\linewidth]{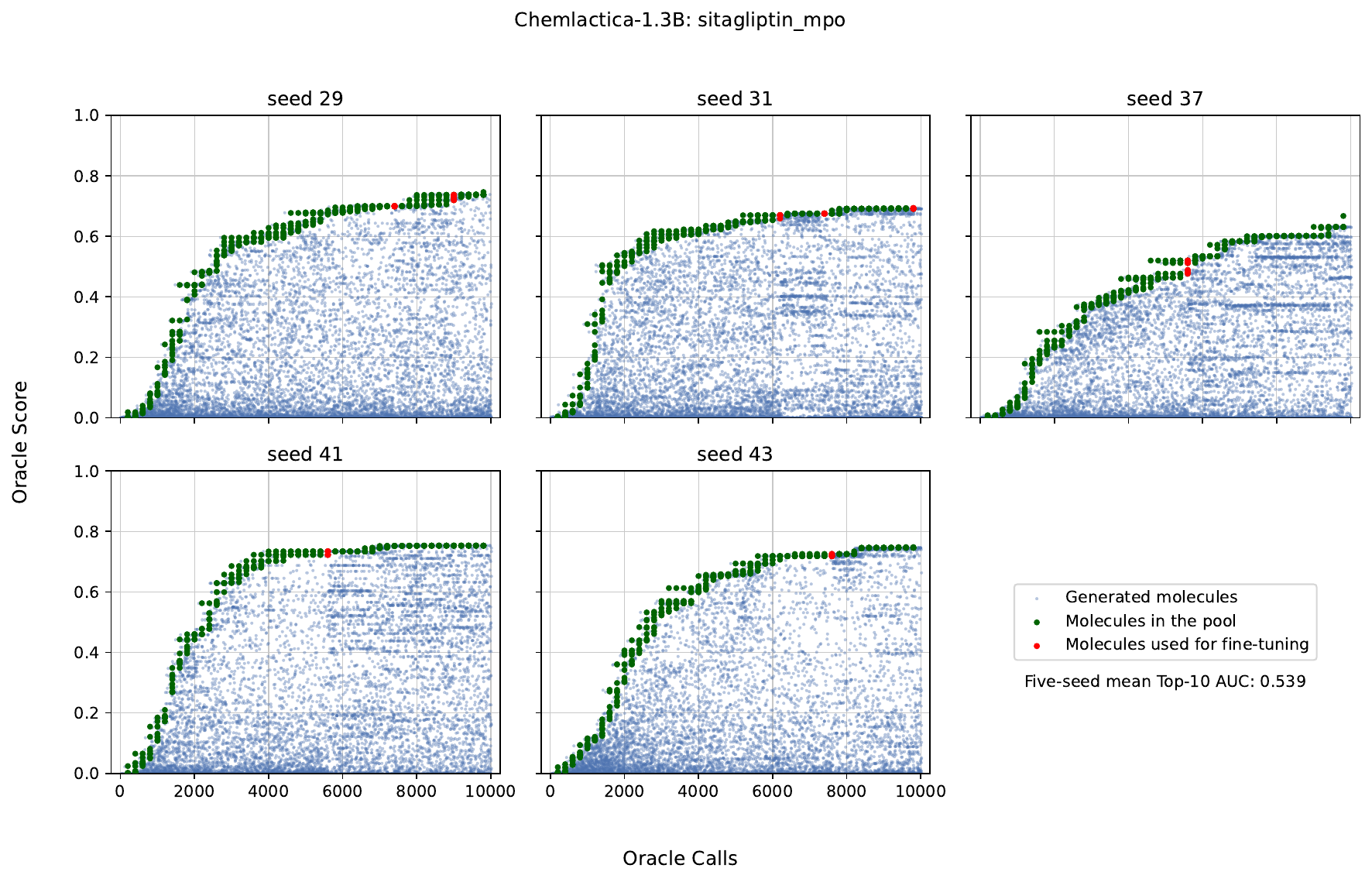}
    \caption{Optimization process visualization using Chemlactica-1.3B for the \texttt{sitagliptin\_mpo} task, using the five independent reproduction seeds.}
    \label{fig:Chemlactica-1.3B-sitagliptin_mpo}
\end{figure}

\begin{figure}
    \centering\includegraphics[width=0.85\linewidth]{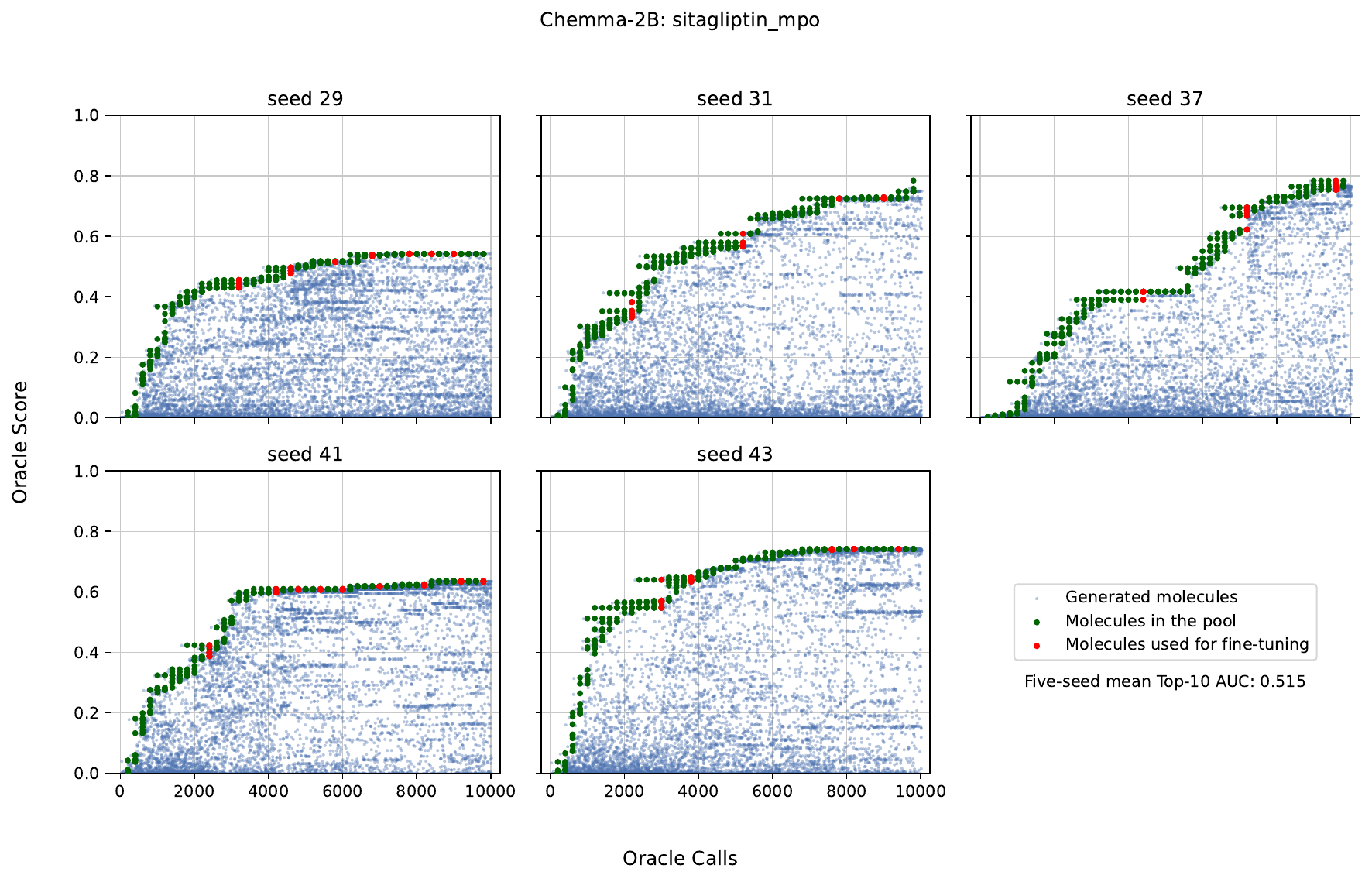}
    \caption{Optimization process visualization using Chemma-2B for the \texttt{sitagliptin\_mpo} task, using the five independent reproduction seeds.}
    \label{fig:Chemma-2B-sitagliptin_mpo}
\end{figure}

\rev{In the task-agnostic setting, Mol-E with Chemlactica-125M reaches a 23-task Top-10 AUC sum of 16.945, exceeding the strongest comparable external result, NovoMolGen at 16.700.} Additionally, strengthening the underlying language model improves performance in the task-agnostic setting: the aggregate score increases to 17.169 with Chemlactica-1.3B and 17.500 with Chemma-2B. Note, however, that Chemlactica and Chemma differ in architecture and pretraining, so this comparison therefore does not constitute a controlled scaling law. In the task-informed setting without target-oracle prescreening, the three Mol-E variants span aggregate scores from 20.192 to 20.551, with no monotonic relationship between parameter count and aggregate performance. \rev{A controlled four-task ablation further shows that the contribution of online fine-tuning is task- and backbone-dependent rather than uniformly positive (Supplementary Section \ref{sec:no-ft-ablation}).} \rev{At the increasingly studied 1,000-call budget, task-informed SEISMO scores 18.156; the strongest Mol-E configuration scores 16.469 and ranks second among comparable results covering all 23 tasks without uncounted target-oracle prescreening (Supplementary Tables \ref{tab:PMO_1k_external} and \ref{tab:PMO_1k_mole}).} Figures \ref{fig:Chemlactica-125M-sitagliptin_mpo}--\ref{fig:Chemma-2B-sitagliptin_mpo} visualize the task-agnostic optimization trajectories for \texttt{sitagliptin\_mpo}. \\

\subsubsection{Novelty of the Generated Molecules}

\rev{To examine how closely the generated molecules resemble the training corpus, we plot their oracle scores against the distance to the neighbor returned by the PubChem USearch index. The distance is one minus the Tanimoto similarity.} \\

\begin{figure}[ht]
    \begin{subfigure}[b]{\textwidth}
        \centering
        \includegraphics[width=0.95\linewidth]{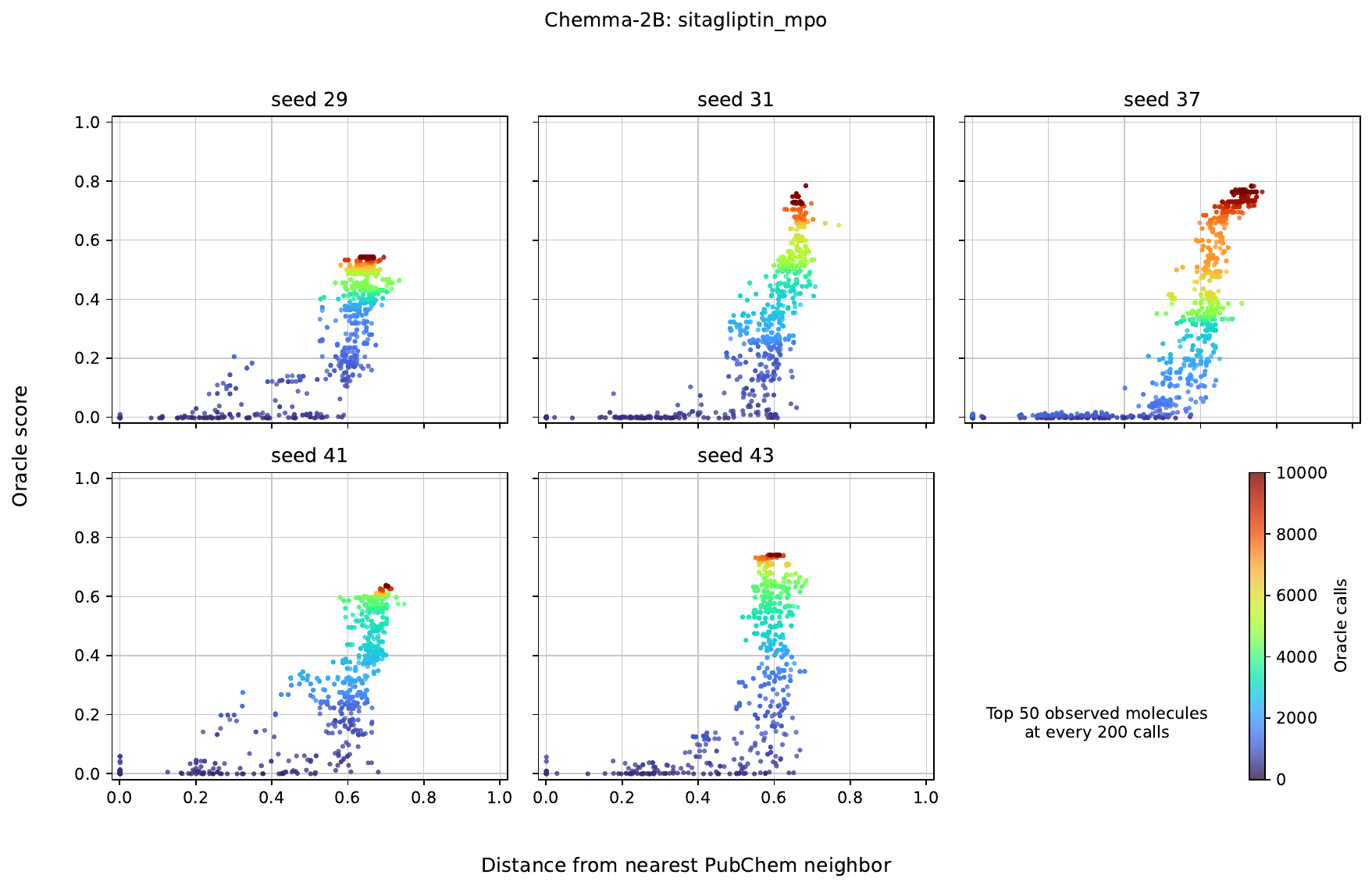}
    \end{subfigure}
    \vspace{0em}
    \begin{subfigure}[b]{\textwidth}
        \centering
        \includegraphics[width=0.95\linewidth]{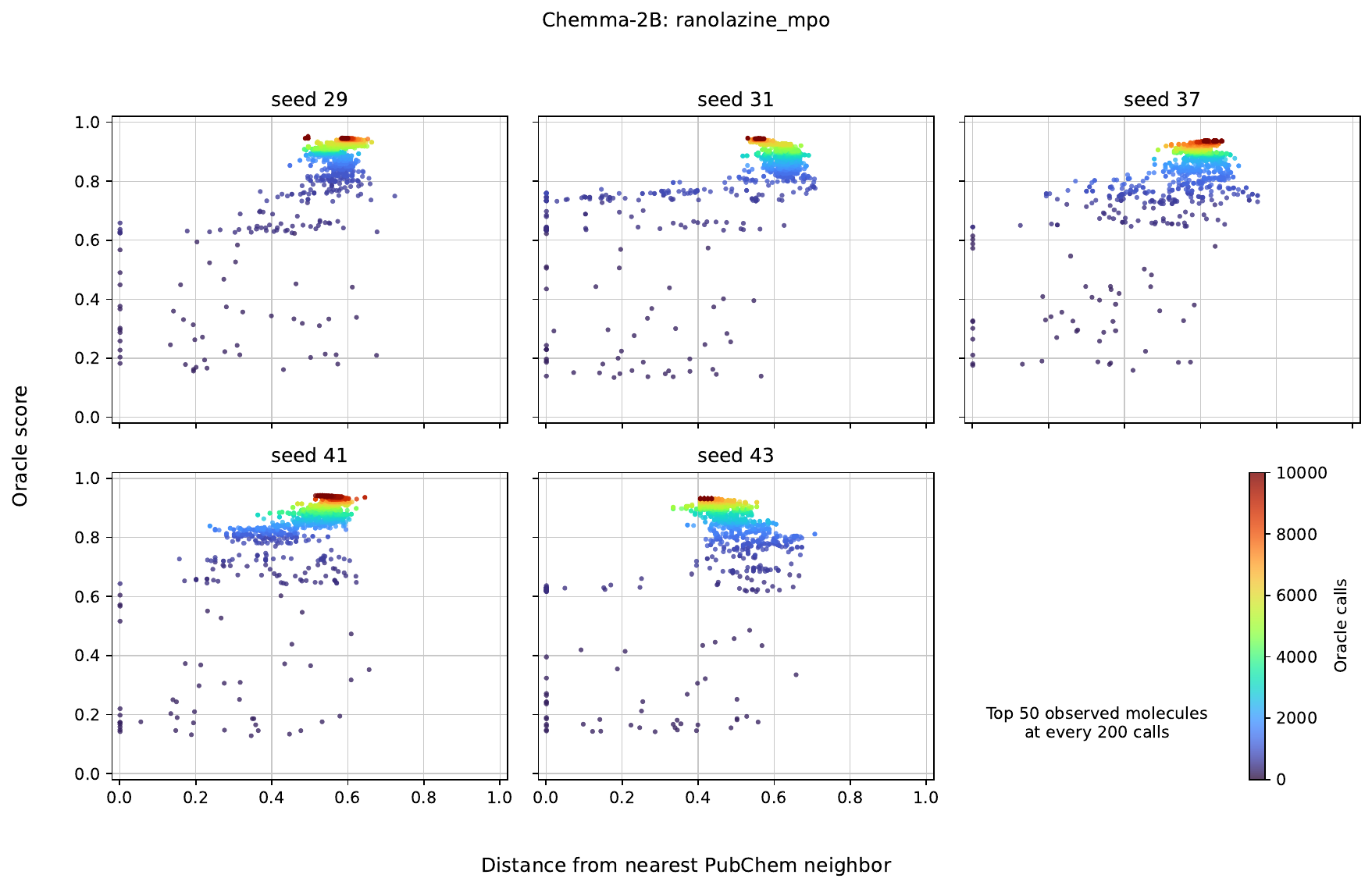}
    \end{subfigure}
    \centering
    \caption{\rev{Oracle score versus distance from the nearest PubChem neighbor returned by USearch for the 50 best molecules observed at every 200-call checkpoint. The top and bottom panels show \texttt{sitagliptin\_mpo} and \texttt{ranolazine\_mpo}, respectively, for the five corrected task-agnostic Chemma-2B seeds. Distance is one minus Tanimoto similarity between radius-2, 2,048-bit Morgan fingerprints.}}
    \label{fig:Chemma-2B-distances}
\end{figure}

\noindent \rev{Figure \ref{fig:Chemma-2B-distances} shows the five corrected task-agnostic Chemma-2B runs for \texttt{sitagliptin\_mpo} and \texttt{ranolazine\_mpo}. At every 200 oracle calls, we plot the 50 highest-scoring molecules observed so far. At the final checkpoint, the seed-level mean nearest-neighbor distances returned by USearch range from 0.610 to 0.726 for Sitagliptin and from 0.418 to 0.622 for Ranolazine, while their corresponding mean oracle scores range from 0.542 to 0.766 and from 0.931 to 0.947, respectively. For Sitagliptin, 48 of the final 50 molecules have nearest-neighbor similarity below 0.3 in two of five seeds, but this threshold is not reached consistently across seeds and is not reached by the final Ranolazine populations. These two examples therefore show that high oracle scores can coexist with moderate-to-large ECFP distance from PubChem neighbors; they do not establish the same behavior on the remaining PMO tasks or exploration beyond the chemical space represented in PubChem.}

\subsection{Multi-property Optimization with Docking}



\subsubsection{Docking MPO on DRD2, MK2, and AChE}
\label{subsec:docking1}

Molecular optimization tasks that incorporate docking simulation serve as valuable benchmarks, since they simulate the real-world drug development setting more closely. Hence, we evaluate Mol-E on the docking-based MPO problems introduced in \cite{augmented-memory}, which focuses on the dopamine type 2 receptor (DRD2), MK2-kinase (MK2), and acetylcholinesterase (AChE) targets. To ensure the generation of realistic molecules, the oracle reward function incorporates the maximization of QED and a molecular weight limit of 500 Da. \\

\noindent Consistent with prior works, we compute the oracle burden and generative yield metrics. Oracle burden measures the number of oracle calls required to generate a specified number of unique molecules above a predefined threshold, and generative yield measures the number of unique molecules generated above a given threshold for a fixed number of oracle calls. To maintain consistency in implementations, we adopt the molecular preprocessing, conformational generation, docking simulation, and reward function from \citep{beam-enumeration}. We reuse the hyperparameters selected for the task-agnostic PMO method in our evolutionary algorithm. \\

\noindent Table \ref{tab:docking} presents our algorithm's performance on this benchmark. Mol-E achieves a significant performance improvement over the Augmented Memory baselines \citep{beam-enumeration}. However, the results indicate that no single model choice in the Mol-E framework dominates the others. Chemlactica-1.3B generally demonstrates superior oracle burden, except for the DRD2 thresholds requiring 10 and 100 qualifying molecules, where Chemlactica-125M is superior. We present our generated structures for all receptors throughout the optimization process in Supplementary Section \ref{sec:docking-molecules-examples}.

\begin{table}
\centering
\resizebox{\columnwidth}{!}{
\begin{tabular}{c|c|ccc|ccc}
\toprule
\multirow{4}{*}{Metric} & \multirow{4}{*}{Target} & \multicolumn{3}{c|}{Augmented Memory \citep{beam-enumeration}} & \multicolumn{3}{c}{Mol-E} \\ \cmidrule{3-8}
& & \multirow{2}{*}{Baseline} & Beam & Beam & Chemlactica & Chemlactica & Chemma \\
 & & & Scaffold 15 & Structure 15 & 125M & 1.3B & 2B \\ \midrule
\multirow{3}{*}{Generative Yield$_{>0.7}$ ($\uparrow$)} & DRD2 & $2513 \pm 442$ & $3412 \pm 95$ & 3474 $\pm$ 158 & 3733 $\pm$ 512 & 3659 $\pm$ 288 & \textbf{3848 $\pm$ 98} \\
 & MK2 & $1446 \pm 173$ & $2584 \pm 443$ & 3127 $\pm$ 138 & \textbf{3772 $\pm$ 578} & 3660 $\pm$ 535 & 3578 $\pm$ 452 \\
 & AChE & $3288 \pm 85$ & $3902 \pm 189$ & 3824 $\pm$ 162 & 4108 $\pm$ 67 & \textbf{4193 $\pm$ 128} & 4092 $\pm$ 284 \\ \midrule
\multirow{3}{*}{Generative Yield$_{>0.8}$ ($\uparrow$)} & DRD2 & $363 \pm 195$ & $1607 \pm 379$ & 1780 $\pm$ 439 & 2827 $\pm$ 510 & 2621 $\pm$ 614 & \textbf{2985 $\pm$ 194} \\
 & MK2 & $34 \pm 13$ & $523 \pm 438$ & 987 $\pm$ 211 & \textbf{2569 $\pm$ 1156} & 2216 $\pm$ 522 & 1058 $\pm$ 465 \\
 & AChE & $556 \pm 47$ & $2124 \pm 326$ & 2059 $\pm$ 327 & 3246 $\pm$ 168 & \textbf{3652 $\pm$ 349} & 3096 $\pm$ 372 \\ \midrule
\multirow{3}{*}{Oracle burden$_{>0.8}$(1) ($\downarrow$)} & DRD2 & $187 \pm 51$ & $83 \pm 29$ & 126 $\pm$ 90 & 20 $\pm$ 29 & \textbf{11 $\pm$ 10} & 74 $\pm$ 62 \\
 & MK2 & $1360 \pm 543$ & $1221 \pm 564$ & 736 $\pm$ 166 & 345 $\pm$ 312 & \textbf{78 $\pm$ 125} & 189 $\pm$ 278 \\
 & AChE & $62 \pm 0$ & $63 \pm 0$ & 105 $\pm$ 29 & 22 $\pm$ 28 & \textbf{15 $\pm$ 23} & 74 $\pm$ 72 \\ \midrule
\multirow{3}{*}{Oracle burden$_{>0.8}$(10) ($\downarrow$)} & DRD2 & $711 \pm 120$ & $571 \pm 104$ & 582 $\pm$ 83 & \textbf{114 $\pm$ 8} & 160 $\pm$ 130 & 240 $\pm$ 11 \\
 & MK2 & $3833 \pm 394$ & $2426 \pm 1525$ & $1122 \pm 154$ & 493 $\pm$ 418 & \textbf{248 $\pm$ 261} & 440 $\pm$ 548 \\
 & AChE & $380 \pm 0$ & $418 \pm 27$ & 462 $\pm$ 25 & 224 $\pm$ 17 & \textbf{91 $\pm$ 103} & 168 $\pm$ 94 \\ \midrule
\multirow{3}{*}{Oracle burden$_{>0.8}$(100) ($\downarrow$)} & DRD2 & $2558 \pm 30$ & $1056 \pm 146$ & 1120 $\pm$ 194 & \textbf{364 $\pm$ 119} & 430 $\pm$ 250 & 518 $\pm$ 41 \\
 & MK2 & Failed & $2676 \pm 403$ & 2189 $\pm$ 181 & 865 $\pm$ 533 & \textbf{486 $\pm$ 346} & 934 $\pm$ 918 \\
 & AChE & $2021 \pm 89$ & $884 \pm 162$ & 1110 $\pm$ 265 & 497 $\pm$ 58 & \textbf{333 $\pm$ 131} & 433 $\pm$ 143 \\ \bottomrule
\end{tabular}}
\caption{Docking MPO experiments run with a maximum budget of 5000 oracle calls. The values represent the mean and the standard deviation of the metrics. The numbers next to the metric names correspond to the thresholds, and the values in parentheses for oracle burden indicate how many unique molecules need to be generated. The best performance on each task-metric combination is bolded.}
\label{tab:docking}
\end{table}

\subsection{Discussion}
Our findings validate the effectiveness of our approach, demonstrating that our models can leverage continued pretraining information to optimize complex reward functions. Furthermore, transferring the optimization hyperparameters selected for the PMO method to the docking MPO tasks in \ref{subsec:docking1} underscores the framework's generalizability. This adaptability suggests that our approach could be particularly valuable when extensive hyperparameter tuning is inaccessible. Specifically, the results demonstrate the applicability of our method to MPO problems with costly oracles, which are more prevalent in industrial drug discovery settings.


\begin{table}
\centering
\resizebox{\columnwidth}{!}{%
\begin{tabular}{@{}lcccccc@{}}
\toprule
\multicolumn{1}{c}{} & \multicolumn{2}{c}{QED}   & \multicolumn{2}{c}{SIM}  & \multicolumn{2}{c}{SAS}    \\ 
\multicolumn{1}{c}{} & PP      & CG              & PP     & CG              & PP      & CG               \\\midrule
Random Init-125M     & 0.030   & 0.110 (0.119)   &  0.068 &  0.239           & 0.092   & 0.442 (0.736)    \\
Chemlactica-125M     & 0.016   & 0.084 (0.084)   &  0.046 &  0.181           & 0.082   & 0.432 (0.432)    \\
Chemlactica-1.3B     & 0.004   & 0.069 (0.096)   &  \textbf{0.043} &    0.172        & 0.064   & 0.281 (0.281)    \\
Chemma-2B-2.1B     & 0.015 & 0.095 (0.095)  & 0.049   &  0.147 &   0.057         & 0.482 (0.482)     \\
Chemma-2B-39B        & \textbf{0.003}   & \textbf{0.059 (0.059)}   &  0.045  &  \textbf{0.167}           & \textbf{0.049}   & \textbf{0.277 (0.277)}    \\\midrule\\ \midrule
                     & \multicolumn{2}{c}{CLOGP} & \multicolumn{2}{c}{TPSA} & \multicolumn{2}{c}{WEIGHT} \\
                     & PP      & CG              & PP     & CG              & PP      & CG               \\\midrule
Random Init-125M     & 0.543   & 1.096 (1.096)   &  2.415 &  9.628 (10.218)           & 11.934   & 58.091 (58.091)    \\
Chemlactica-125M     & 0.101   & 0.429 (0.482)   & 1.326  & 7.876 (8.774)   & 6.996   & 17.267 (17.267)  \\
Chemlactica-1.3B     & 0.094   & 0.507 (0.507)   & 1.032  & \textbf{5.579 (5.579)}   & 5.699   & \textbf{13.494 (13.679)}    \\
    Chemma-2B-2.1B       & 0.055 & 0.700 (0.700) & 1.662 & 6.307 (6.307) &4.187 & 19.565 (19.565)\\
Chemma-2B-39B        & \textbf{0.037}   & \textbf{0.454 (0.454)}   & \textbf{0.933}  & 7.091 (7.091)   & \textbf{0.640}   & 15.429 (15.429)  \\ \bottomrule
\end{tabular}}
\caption{RMSE (RMSE corrected for the mean) $\downarrow$ for Property Prediction (PP) and Conditional Generation (CG) for different tasks and models.}
\label{tab:cg-pp}
\end{table}

\subsection{\rev{Fine-Tuned ADME Property Prediction}}

\rev{We evaluate supervised fine-tuning on the six assay-derived ADME endpoints released by Fang et al. \citep{admet-industrial}. Table \ref{tab:adme} compares the three molecular language models with published models evaluated on the same official Polaris single-task train--test splits.}

\begin{table}[t]
\centering
\resizebox{\columnwidth}{!}{%
\begin{tabular}{lcccccc}
\toprule
\rev{Model} & \rev{HLM} & \rev{MDR} & \rev{SOL} & \rev{RLM} & \rev{HPPB} & \rev{RPPB} \\ \midrule
\rev{CheMeleon \citep{burns2026chemeleon}} & \rev{0.720} & \rev{0.822} & \rev{0.682} & \rev{0.757} & \rev{0.793} & \rev{0.662} \\
\rev{Fang et al. Random Forest \citep{admet-industrial}} & \rev{0.639$^{a}$} & \rev{0.716$^{a}$} & \rev{0.439$^{b}$} & \rev{0.640$^{a}$} & \rev{0.690$^{c}$} & \rev{0.722$^{a}$} \\
\rev{MolGPS: 1B MPNN (LargeMix + phenomics) \citep{sypetkowski2024molgps}} & \rev{\textbf{0.778}} & \rev{\textbf{0.860}} & \rev{\textbf{0.764}} & \rev{\textbf{0.784}} & \rev{\textbf{0.888}} & \rev{\textbf{0.908}} \\
\rev{MolEncoder \citep{kruger2025molencoder}} & \rev{0.714} & \rev{0.804} & \rev{0.631} & \rev{--} & \rev{--} & \rev{--} \\ \midrule
\rev{Chemlactica-125M} & \rev{0.717} & \rev{0.714} & \rev{0.608} & \rev{0.714} & \rev{0.774} & \rev{0.442} \\
\rev{Chemlactica-1.3B} & \rev{0.720} & \rev{0.762} & \rev{0.574} & \rev{0.698} & \rev{0.635} & \rev{0.614} \\
\rev{Chemma-2B} & \rev{0.674} & \rev{0.709} & \rev{0.558} & \rev{0.660} & \rev{0.636} & \rev{0.747} \\ \bottomrule
\end{tabular}}
\caption{\rev{Pearson correlation $\uparrow$ on the held-out test sets of the six official Polaris/Fang single-task ADME benchmarks. Values are held-out test submissions. HLM and RLM denote human and rat liver microsomal stability; MDR denotes MDR1--MDCK efflux ratio; SOL denotes solubility; and HPPB and RPPB denote human and rat plasma protein binding. The MolGPS-family row is the source paper's 1B MPNN submission trained on LargeMix and phenomics. MolEncoder reports results for only HLM, MDR, and SOL. Random Forest features are $^{a}$2D descriptors, $^{b}$FCFP4, and $^{c}$atom-pair fingerprints.}}
\label{tab:adme}
\end{table}

\rev{The specialist graph models generally obtain higher correlations than our autoregressive models. Across our three models, the strongest result depends on the endpoint: Chemlactica-1.3B performs best on HLM and MDR, Chemlactica-125M on SOL, RLM, and HPPB, and Chemma-2B on RPPB.}

\subsection{Limitations}
The language models introduced in this paper operate only on SMILES representations and do not support 3D coordinates of atoms, limiting their reliability in scenarios where 3D conformation is critical. Furthermore, the models have a very limited understanding of biological entities like proteins, constraining their practical applicability in particular biochemistry and drug discovery use-cases. While effective, the optimization algorithms presented in this paper are not exhaustively tuned, suggesting potential room for improvement. Finally, our current approach does not directly account for or consider synthetic accessibility or other practical considerations in drug design, which limits its immediate applicability in real-world drug discovery pipelines.

\section{Conclusion}
This paper presents three language models: Chemlactica-125M, Chemlactica-1.3B, and Chemma-2B, trained on a novel corpus encompassing over 100 million molecules and their properties. We demonstrate the efficacy of these models on multiple tasks inspired by industrial drug design, with a particular focus on molecular optimization. Our proposed optimization algorithm combines the capabilities of language models with concepts from genetic algorithms. \rev{On PMO-10K, Mol-E obtains the highest aggregate Top-10 AUC among the comparable full-23-task results considered in both the task-agnostic and task-informed regimes.}
We publicly release our training corpus, pretrained models, optimization algorithm, and associated training recipes to support reproducibility, making an early step toward applying language models to chemical research. We hope our contributions provide a valuable foundation for future work in this domain, enabling new approaches for molecular design and analysis.

\section*{Abbreviations}
\noindent AChE, acetylcholinesterase; ADME, absorption, distribution, metabolism, and excretion; AUC, area under the curve; CLogP, calculated octanol--water partition coefficient; DRD2, dopamine D2 receptor; ECFP, extended-connectivity fingerprint; LLM, large language model; MK2, MAPK-activated protein kinase 2; Mol-E, Molecular Language Model Powered Evolutionary Algorithm; MPO, multi-property optimization; PMO, Practical Molecular Optimization; QED, quantitative estimate of drug-likeness; RMSE, root mean square error; SAS, synthetic accessibility score; SMILES, simplified molecular-input line-entry system; TPSA, topological polar surface area.

\section*{Declarations}

\bmhead{Availability of data and materials}
Source code and data-preparation resources are available under the MIT License at \url{https://github.com/yerevann/chemlactica}. The PubChemForLM training corpus is publicly available at \url{https://huggingface.co/datasets/yerevann/PubChemForLM}. The pretrained Chemlactica-125M, Chemlactica-1.3B, and Chemma-2B model checkpoints are available at \url{https://huggingface.co/yerevann/chemlactica-125m}, \url{https://huggingface.co/yerevann/chemlactica-1.3b}, and \url{https://huggingface.co/yerevann/chemma-2b}, respectively.

\bmhead{Competing interests}
The authors declare no competing interests.

\bmhead{Funding}
This research was supported by the Higher Education and Science Committee of MESCS RA (Research project No. 24FP-1A058). This work was also supported by Nebius.ai through access to GPU cloud resources and technical support, and by Yandex Armenia through a fellowship awarded to Philipp Guevorguian. The funders had no role in the study design, data collection and analysis, decision to publish, or preparation of the manuscript.

\bmhead{Authors' contributions}
Philipp Guevorguian led model pretraining and data collection and wrote the initial manuscript. Tigran Fahradyan designed the molecular optimization algorithm and conducted the PMO and docking experiments. Menua Bedrosian conducted the property-prediction experiments and evaluated conditional generation. Gayane Chilingaryan contributed to pretraining data collection. Hrant Khachatrian supervised the project and contributed to the manuscript's framing and writing. Armen Aghajanyan supervised model pretraining and fine-tuning. All authors reviewed and approved the final manuscript.

\bmhead{Acknowledgements}
We thank Garik Petrosyan and Zaven Navoyan for insightful discussions.


\section*{Supplementary Information} 
\setcounter{section}{1} 
\renewcommand{\thesection}{S\arabic{section}} 
\renewcommand{\thesubsection}{S\arabic{section}.\arabic{subsection}} 

\subsection{Dataset Generation}
\label{sec:data_added}
\subsubsection{Data Collection}
We first constructed a comprehensive SQL database using PubChem dumps to generate our training corpus. Then, using RDKit \citep{RDKIT}, we computed key molecular properties, including synthetic accessibility score (SAS), quantitative estimate of drug-likeness (QED), molecular weight (MW), total polar surface area (TPSA), partition coefficient (CLogP), and various structural features such as hydrogen donors/acceptors and ring counts. Due to differences in SMILES canonicalization between PubChem and RDKit, we standardized all SMILES strings using RDKit's implementation. \\

\noindent Our dataset's cutoff date is January 26, 2023, thus excluding any subsequent additions or modifications to PubChem. To ensure data integrity, molecules that failed RDKit's MolFromSmiles parsing were discarded. To incorporate similarity information, we utilized PubChem's related molecule data, which includes pairs with Tanimoto similarity $\ge$0.8 based on PubChem fingerprints. From the resulting 200 billion pairs, we sampled ~4 billion and recalculated their similarities using the ECFP4 fingerprint for improved accuracy and consistency with other methods.

\subsubsection{JSONL Corpus Generation} 
We transformed our database into a corpus of JSONL files, with each molecule represented as a single JSON object. This representation includes molecular identifiers, computed properties, similarity data, synonyms, experimental properties, and the PubChem compound identifier (CID). This representation allows for more flexible manipulation of molecules' text representation, which we describe in \ref{sec:data_main}.

\subsection{Training Details}
\label{sec:training_details}

\subsubsection{Tokenization}
\label{subsec:tokenization}
We utilized the original tokenizers from Gemma and Galactica, adding chemistry-specific tokens \prompt{[START\_SMILES]} and \prompt{[END\_SMILES]} to Gemma's tokenizer for consistency. To optimize training efficiency, we included all opening and closing tags as special tokens (e.g., \prompt{[QED]}). Samples of varying lengths were tokenized and grouped into blocks of 2048 tokens, separated by model-specific separator tokens (EOS \prompt{</s>} for Chemlactica, BOS \prompt{<bos>} for Chemma).

\subsubsection{Training Implementation}
Chemma and Chemlactica were trained using the AdamW optimizer \citep{AdamW} with cross-entropy loss and a causal language modeling objective. We applied dropout only to Chemlactica, maintaining consistency with the original model training regimes. For computational efficiency, we train Chemma-2B in full bfloat16. We leveraged PyTorch's \citep{pytorch} Fully Sharded Data Parallel (FSDP) \citep{fsdp} and Flash Attention \citep{flash-attention-2} for optimized training. The training was conducted locally (Chemlactica-125M: 306 A100 hours) and on Nebius.ai cloud (Chemma-2B: 488 H100 GPU hours, Chemlactica-1.3B: 288 H100 GPU hours). Preparatory work before the final training runs consumed multiple thousands of A100 hours.

\subsubsection{Model Training Hyperparameters} 
Table \ref{tab:hyp} lists the hyperparameters we used for pretraining the aforementioned models.

\begin{table}[b]
\centering
\begin{tabular}{lrrr}
\toprule
             & Chemlactica-125M         &  Chemlactica-1.3B&Chemma-2B                \\\midrule
Peak learning rate& 1.4e-3                     & 1.0e-4                     & 1.0e-3                     \\
Warmup steps      & 500                     & 500                     & 500                     \\
Context length    & 2048                     & 2048                     & 2048                     \\
ADAM $\beta_1$       & 0.9                      &  0.9&0.9                      \\
ADAM $\beta_2$       & 0.95                     &  0.95                     &0.95                     \\
ADAM $\epsilon$      & 1e-8                     &  1e-8                     &1e-8                     \\
Weight Decay      & 0.1                      &  0.1                      &0.1                      \\
Dropout           & 0.1                      &  0.1                      &None                     \\
Attention Dropout & 0.1                      &  0.1                      &None                     \\
Precision         & Mixed                    &  Mixed                    &BF16                     \\
Loss Function     & CE Loss                  &  CE Loss                 &CE Loss                  \\
Vocabulary Size   & 50066                    &  50066                    &256000                   \\
Gradient Clipping & 1.0                      &  1.0                      &1.0                     \\ \bottomrule
\end{tabular}
\caption{Hyperparameters of our language models. All cross-entropy losses use mean reduction.}
\label{tab:hyp}
\end{table}



\subsection{Hyperparameter Tuning and Selection}
\label{sec:hparams}

\subsubsection{Optimization Algorithm Hyperparameter Tuning for PMO.}
\label{subsec:pmo-hparams}
Given our optimization algorithm's large number of hyperparameters, we adopt a two-phase approach. First, we identify and freeze the hyperparameters that empirically show less sensitivity to the algorithm's performance. Then, we focus on tuning the more sensitive hyperparameters using grid search. We tune the hyperparameters separately for Chemlactica-125M, Chemlactica-1.3B, and Chemma-2B to account for model-specific optimal settings. \rev{For approximate nearest-neighbor search in PubChem, we use the USearch vector-search engine \citep{usearch}. The index contains 116,134,874 radius-2, 2,048-bit Morgan fingerprints and uses the Tanimoto metric. We query it at a search expansion of 16,384, repeat the search at an expansion of 4,096 to assess search stability, and verify every returned distance by direct bit-vector Tanimoto computation. For the 500 molecules in the final populations used for the numerical statements in Section \ref{sec:results}, increasing the expansion from 16,384 to 65,536 changes none of the returned neighbors or distances.}\\

\noindent \textit{Selection criteria.} For tuning, we utilize the \texttt{perindopril\_mpo} and \texttt{zaleplon\_mpo} tasks from the PMO benchmark, following the methodology in \citep{pmo}. We report the AUC Top-10 metric from three independent runs with different seeds for each hyperparameter configuration. \\

\noindent \textit{Fixed hyperparameters and grid.} A key hyperparameter, $N$, which determines the number of molecules generated before updating the pool, is set to 200. We implement a dynamic temperature scheduling strategy to increase the diversity of generated molecules. The sampling temperature starts at 1 and linearly increases to 1.5 as the number of oracle evaluations grows. This gradual temperature increase promotes the generation of more diverse molecules over time, reducing repetition and encouraging exploration of the chemical space. \\

\noindent For computational batching, the generation batch sizes are 128, 120, and 32 for Chemlactica-125M, Chemlactica-1.3B, and Chemma-2B, respectively, and are fixed across all PMO tasks, information regimes, and seeds for each model. \\

\noindent We perform grid search on $P$ (pool size), $S$ (number of similar molecules), $K$ (fine-tuning tolerance level), and $lr$ (fine-tuning peak learning rate) with the following grid:

\begin{itemize}
    \item $P = [10, 30, 50]$
    \item $S = [0, 1, 2, 5]$
    \item $K = [3, 5, 7]$
    \item $lr = [10^{-4}, 10^{-5}]$
\end{itemize}

\subsubsection{Conditional Generation Hyperparameters}
\label{sec:CGdetails}
Our generation process benefits from the following techniques to improve output quality:

\begin{itemize}
    \item \textbf{Chain-of-Thought (CoT):} We omit \prompt{[START\_SMILES]} from the initial prompt, enabling the model to generate more property values before the molecule itself.
    \item \textbf{Repetition Penalty:} Applied to discourage repetitive outputs.
    \item \textbf{Undesired Token Suppression:} Employed to ensure the model eventually generates \prompt{[START\_SMILES]}.
\end{itemize}

\noindent Table \ref{tab:ab-cg} provides an ablation study of these sampling components across our three models, demonstrating their individual and combined impacts on generation quality. Surprisingly, the best combination of hyperparameters, as chosen by lowest corrected RMSE (RMSE(c)), coincides with all three models. DNF (Did Not Finish) trial exceeded 30 minutes of runtime when it was manually terminated.

\begin{table}[t]
\resizebox{\columnwidth}{!}{
\begin{tabular}{cccrrrrrr}

\toprule
             &                &              & \multicolumn{2}{c}{Chemlactica-125M}    & \multicolumn{2}{c}{Chemlactica-1.3B}    & \multicolumn{2}{c}{Chemma-2B}           \\
CoT          & rep.           & supp.        & RMSE (c) $\downarrow$             & Invalids $\downarrow$     & RMSE (c) $\downarrow$              & Invalids $\downarrow$      & RMSE (c) $\downarrow$               & Invalids $\downarrow$       \\ \midrule
No           & 1.0            & No           & 70.11 (70.11)          & 0/100          & 15.81 (65.32)          & 1/100          & 12.15 (64.54)          & 1/100          \\
Yes          & 1.0            & No           & 112.52 (112.52)        & 0/100          & 187.26 (187.26)        & 0/100          & 198.48 (191.89)        & 46/100         \\
Yes          & 1.010          & No           & 82.28 (82.28)          & 0/100          & 137.19 (137.19)        & 0/100          & 170.02 (170.02)        & 0/100          \\
Yes          & 1.0            & Yes          & 33.46 (33.46)          & 0/100          & 18.53 (25.22)          & 1/100          & 31.98 (31.85)          & 1/100          \\
Yes          & 1.005          & Yes          & 34.52 (34.52)          & 0/100          & 17.14 (17.14)          & 0/100          & 29.71 (29.71)          & 0/100          \\
\textbf{Yes} & \textbf{1.010} & \textbf{Yes} & \textbf{30.27 (30.27)} & \textbf{0/100} & \textbf{16.87 (16.87)} & \textbf{0/100} & \textbf{18.93 (20.39)} & \textbf{1/100} \\
Yes          & 1.015          & Yes          & 30.27 (30.27)          & 0/100          & 18.07 (19.61)          & 1/100          & 18.99 (20.44)          & 1/100          \\
Yes          & 1.020          & Yes          & 31.17 (31.17)          & 1/100          & 16.33 (18.03)          & 1/100          & 24.16 (25.27)          & 1/100          \\
Yes          & 1.050          & Yes          & 45.38 (45.38)          & 1/100          & 16.49 (34.48)          & 1/100          & 74.78 (130.11)         & 63/100         \\
Yes          & 1.100          & Yes          & 35.20 (35.20)          & 0/100          & 16.61 (32.37)          & 1/100          & 740.28 (488.73)        & 59/100         \\ \midrule

No  & 1.0   & No    & 0.268 (0.769)         & 19/100                & 0.395 (0.395)          & 1/100                 & 0.391 (0.431           & 4/100                 \\
Yes & 1.0   & No    & 0.887 (0.887)         & 0/100                 & 0.866 (0.866)          & 0/100                 & DNF                    & 46/100                \\
Yes & 1.010 & No    & 0.951 (0.951)         & 0/100                 & 0.691 (0.691)          & 0/100                 & 0.769 (0.769)          & 0/100                 \\
Yes & 1.0   & Yes   & 0.436 (0.436)         & 0/100                 & 0.470 (0.470)          & 2/100                 & 0.253 (0.253)          & 1/100                 \\
Yes & 1.005 & Yes   & 0.439 (0.439)         & 0/100                 & 0.475 (0.475)          & 0/100                 & 0.348 (0.363)          & 1/100                 \\
Yes & 1.010 & Yes   & 0.432 (0.432)         & 0/100                 & \textbf{0.281 (0.281)} & \textbf{0/100}        & \textbf{0.275 (0.275)} & \textbf{0/100}        \\
Yes & 1.015 & Yes   & 0.373 (0.378)         & 1/100                 & 0.540 (0.536)          & 2/100                 & 0.331 (0.347)          & 2/100                 \\
Yes & 1.020 & Yes   & 0.432 (0.432)         & 0/100                 & 0.369 (0.369)          & 0/100                 & 0.325 (0.341)          & 2/100                 \\
Yes & 1.050 & Yes   & 0.294 (0.733)         & 3/100                 & 0.843 (0.951)          & 10/100                & DNF                    & 63/100                \\
\textbf{Yes} & \textbf{1.100} & \textbf{Yes}   & \textbf{0.381 (0.381)}         & 0/100                 & 0.449 (0.449)          & 1/100                 & DNF                    & 59/100                \\ \bottomrule
\end{tabular}}
\caption{Ablation study on Conditional Generation hyperparameters. Each row represents one combination of Chain-of-Thought (CoT), repetition penalty (rep.), and suppression (supp.). All experiments are done on the molecular weight (top) and SAS (bottom) prediction tasks.}
\setlength{\tabcolsep}{3pt}
\label{tab:ab-cg}
\end{table}

\subsection{The algorithm for converting molecules to prompt}

Algorithm \ref{alg:molecules2prompt} shows the procedure of converting a molecule and its similar molecule into either a prompt for new molecule generation or a training sample. The separator token represented by \prompt{<sep>} corresponds to the eos token \prompt{</s>} for the Chemlactica models and the bos token \prompt{<bos>} for Chemma. 

\begin{algorithm}[tb]
   \caption{\textbf{molecules2prompt}}
   \label{alg:molecules2prompt}
\begin{algorithmic}
   \State {\bfseries Input:} $m_1, m_2, \ldots, m_S$
   \State 1. Sample similarity values for molecules in the prompt
   \State $v_i^{sim} \sim \mathcal{U}(0.4, 0.9), i=1, \ldots, S$
   \State
   \State 2. Concatenate all molecules with their similarity values
   \State $p \leftarrow$ \prompt{[SIMILAR]$m_{1}^{smiles}$ $v_{1}^{sim}$[/SIMILAR]...[SIMILAR]$m_{S}^{smiles}$ $v_{S}^{sim}$[/SIMILAR]}
   \State
   \State 3. Add a desirable oracle score to the prompt
   \If{at least one fine-tuning has been performed}
   \State $v^{max} \leftarrow$ the maximum oracle score achieved so far
   \State $v^{oracle} \sim \mathcal{U}(v^{max}, oracle\_max)$
   \State $p \leftarrow concat(p,$ \prompt{[PROPERTY]$v^{oracle}$[/PROPERTY]}$)$
   \EndIf
   \State
   \State 4. Add molecule generation suffix
   \State \textbf{return} $concat(p, $\prompt{[START\_SMILES]}$)$
   
\end{algorithmic}
\end{algorithm}

\begin{algorithm}[tb]
   \caption{\textbf{molecules2sample}}
   \label{alg:molecules2sample}
\begin{algorithmic}
   \State {\bfseries Input:} $(m_1, m_2, \ldots, m_S), m$
   \State 1. Compute the similarity values for the molecules in the training sample
   \State $v_i^{sim} = similarity(m_i, m), i=1, \ldots, S$
   \State
   \State 2. Concatenate all molecules with their similarity values
   \State $p \leftarrow$ \prompt{[SIMILAR]$m_{1}^{smiles}$ $v_{1}^{sim}$[/SIMILAR]...[SIMILAR]$m_{S}^{smiles}$ $v_{S}^{sim}$[/SIMILAR]}
   \State
   \State 3. Add the oracle score to the training sample
   \If{at least one fine-tuning has been performed}
   \State $v^{oracle}$ = $O(m)$
   \State $p \leftarrow concat(p,$ \prompt{[PROPERTY]$v^{oracle}$[/PROPERTY]}$)$
   \EndIf
   \State
   \State 4. Add training sample suffix
   \State \textbf{return} $concat(p, $\prompt{[START\_SMILES]$m^{smiles}$[END\_SMILES]<sep>}$)$
   
\end{algorithmic}
\end{algorithm}

\clearpage
\subsection{Detailed Results for Practical Molecular Optimization}
\label{sec:pmo-detailed-results}

\rev{Tables \ref{tab:PMO_full} and \ref{tab:PMO_full_recent_agnostic} report task-agnostic PMO-10K results: the first contains the original PMO baselines and Mol-E reproductions, while the second collects additional recent methods in a single comparison. Table \ref{tab:PMO_full_task_informed} reports task-informed results, and Table \ref{tab:PMO_full_budget_noncomparable} separately identifies methods with uncounted target-oracle use. Tables \ref{tab:PMO_1k_external} and \ref{tab:PMO_1k_mole} give the corresponding PMO-1K landscape. Unreported Valsartan AUC values in otherwise complete PMO-10K evaluations are conservatively assigned zero; changed oracle implementations and omitted oracle work are identified explicitly.} The binary multiplier in the \texttt{valsartan\_smarts} oracle provides no signal for most evaluated molecules, so few task-agnostic trajectories escape the zero-reward region, whereas task-informed prompting substantially improves this task.

\begin{table}[!h]
\centering
\resizebox{\columnwidth}{!}{
\begin{tabular}{cccc|ccc}
\toprule
Oracle                    & REINVENT                   & Augmented                  & Genetic                    & Chemlactica                            & Chemlactica                            & Chemma                                          \\
                          &                            & Memory                     & GFN                        & 125M                                   & 1.3B                                   & 2B                                              \\ \midrule
albuterol\_similarity     & 0.882 $\pm$ 0.006          & 0.913 $\pm$ 0.009          & 0.949 $\pm$ 0.010          & 0.970 $\pm$ 0.004 & 0.951 $\pm$ 0.008 & 0.959 $\pm$ 0.008 \\
amlodipine\_mpo           & 0.635 $\pm$ 0.035          & 0.691 $\pm$ 0.047          & 0.761 $\pm$ 0.019          & 0.760 $\pm$ 0.093 & 0.821 $\pm$ 0.049 & 0.759 $\pm$ 0.100 \\
celecoxib\_rediscovery & 0.713 $\pm$ 0.067          & 0.796 $\pm$ 0.008          & 0.802 $\pm$ 0.029          & 0.830 $\pm$ 0.122 & 0.903 $\pm$ 0.014 & 0.918 $\pm$ 0.015 \\
deco\_hop                 & 0.666 $\pm$ 0.044          & 0.658 $\pm$ 0.024          & 0.733 $\pm$ 0.109          & 0.820 $\pm$ 0.109 & 0.788 $\pm$ 0.145 & 0.830 $\pm$ 0.120 \\
drd2                      & 0.945 $\pm$ 0.007          & 0.963 $\pm$ 0.006          & 0.974 $\pm$ 0.006 & \textbf{0.979 $\pm$ 0.003} & 0.976 $\pm$ 0.005 & 0.975 $\pm$ 0.007 \\
fexofenadine\_mpo         & 0.784 $\pm$ 0.006          & 0.859 $\pm$ 0.009          & 0.856 $\pm$ 0.039          & 0.917 $\pm$ 0.028 & 0.909 $\pm$ 0.032 & 0.935 $\pm$ 0.014 \\
gsk3b                     & 0.865 $\pm$ 0.043          & 0.881 $\pm$ 0.021          & 0.881 $\pm$ 0.042          & 0.939 $\pm$ 0.016 & 0.947 $\pm$ 0.014 & 0.943 $\pm$ 0.016 \\
isomers\_c7h8n2o2         & 0.852 $\pm$ 0.036          & 0.853 $\pm$ 0.087          & 0.969 $\pm$ 0.003          & 0.903 $\pm$ 0.036 & 0.922 $\pm$ 0.016 & 0.924 $\pm$ 0.008 \\
isomers\_c9h10n2o2pf2cl   & 0.642 $\pm$ 0.054          & 0.736 $\pm$ 0.051          & 0.897 $\pm$ 0.007          & 0.837 $\pm$ 0.041 & 0.816 $\pm$ 0.049 & 0.852 $\pm$ 0.043 \\
jnk3                      & 0.783 $\pm$ 0.023          & 0.739 $\pm$ 0.110          & 0.764 $\pm$ 0.069          & 0.877 $\pm$ 0.047 & 0.854 $\pm$ 0.114 & 0.914 $\pm$ 0.023 \\
median1                   & 0.356 $\pm$ 0.009          & 0.326 $\pm$ 0.013          & 0.379 $\pm$ 0.010          & 0.359 $\pm$ 0.030 & 0.361 $\pm$ 0.009 & 0.377 $\pm$ 0.009 \\
median2                   & 0.276 $\pm$ 0.008          & 0.291 $\pm$ 0.008          & 0.294 $\pm$ 0.007          & 0.336 $\pm$ 0.015 & 0.354 $\pm$ 0.022 & 0.358 $\pm$ 0.028 \\
mestranol\_similarity     & 0.618 $\pm$ 0.048          & 0.750 $\pm$ 0.049          & 0.708 $\pm$ 0.057          & 0.925 $\pm$ 0.048 & 0.890 $\pm$ 0.076 & 0.911 $\pm$ 0.027 \\
osimertinib\_mpo          & 0.837 $\pm$ 0.009          & 0.855 $\pm$ 0.004          & 0.860 $\pm$ 0.008          & 0.899 $\pm$ 0.013 & 0.909 $\pm$ 0.019 & 0.891 $\pm$ 0.011 \\
perindopril\_mpo          & 0.537 $\pm$ 0.016          & 0.613 $\pm$ 0.015          & 0.595 $\pm$ 0.014          & 0.651 $\pm$ 0.037 & 0.755 $\pm$ 0.024 & 0.724 $\pm$ 0.114 \\
qed                       & 0.941 $\pm$ 0.000          & \textbf{0.942 $\pm$ 0.000} & \textbf{0.942 $\pm$ 0.000} & \textbf{0.942 $\pm$ 0.001} & \textbf{0.942 $\pm$ 0.000} & \textbf{0.942 $\pm$ 0.000} \\
ranolazine\_mpo           & 0.760 $\pm$ 0.009          & 0.801 $\pm$ 0.006          & 0.819 $\pm$ 0.018          & 0.885 $\pm$ 0.021 & 0.895 $\pm$ 0.018 & 0.890 $\pm$ 0.007 \\
scaffold\_hop             & 0.560 $\pm$ 0.019          & 0.567 $\pm$ 0.008          & 0.615 $\pm$ 0.100          & 0.668 $\pm$ 0.082 & 0.617 $\pm$ 0.032 & 0.709 $\pm$ 0.125 \\
sitagliptin\_mpo          & 0.021 $\pm$ 0.003          & 0.284 $\pm$ 0.050          & 0.634 $\pm$ 0.039          & 0.603 $\pm$ 0.048 & 0.539 $\pm$ 0.062 & 0.515 $\pm$ 0.062 \\
thiothixene\_rediscovery  & 0.534 $\pm$ 0.013          & 0.550 $\pm$ 0.041          & 0.583 $\pm$ 0.034          & 0.682 $\pm$ 0.079 & 0.678 $\pm$ 0.085 & 0.655 $\pm$ 0.130 \\
troglitazone\_rediscovery & 0.441 $\pm$ 0.032          & 0.540 $\pm$ 0.048          & 0.511 $\pm$ 0.054          & 0.725 $\pm$ 0.151 & 0.781 $\pm$ 0.148 & 0.877 $\pm$ 0.013 \\
valsartan\_smarts         & \textbf{0.178 $\pm$ 0.358} & 0.000 $\pm$ 0.000          & 0.135 $\pm$ 0.271          & 0.000 $\pm$ 0.000 & 0.161 $\pm$ 0.322 & 0.177 $\pm$ 0.354 \\
zaleplon\_mpo             & 0.358 $\pm$ 0.062          & 0.394 $\pm$ 0.026          & 0.552 $\pm$ 0.033          & 0.437 $\pm$ 0.014 & 0.397 $\pm$ 0.047 & 0.464 $\pm$ 0.043 \\ \midrule
sum                       & 14.196                     & 15.002                     & 16.213                     & 16.945 $\pm$ 0.237 & 17.169 $\pm$ 0.626 & \textbf{17.500 $\pm$ 0.840} \\ \bottomrule
\end{tabular}}
\caption{Task-agnostic comparison on all PMO tasks for methods with available per-task uncertainty estimates. Values are mean Top-10 AUC $\uparrow$ $\pm$ population standard deviation over five independent runs; the Mol-E columns use only the corrected reproduction results.}
\label{tab:PMO_full}
\end{table}

\begin{table}[!h]
\begingroup
\centering
\scriptsize
\setlength{\tabcolsep}{2pt}
\resizebox{\columnwidth}{!}{
{\color{black}
\begin{tabular}{l*{12}{c}}
\toprule
Oracle & \rotatebox{90}{LICO} & \rotatebox{90}{GP-MoLFormer Sim+GA} & \rotatebox{90}{GPBO} & \rotatebox{90}{ACEGEN PPOD} & \rotatebox{90}{InVirtuoGen cold start} & \rotatebox{90}{SEGO} & \rotatebox{90}{SEISMO no description} & \rotatebox{90}{NovoMolGen} & \rotatebox{90}{ChemTSv3$^{\dagger}$} & \rotatebox{90}{SynGBO} & \rotatebox{90}{S3-GFN} & \rotatebox{90}{GenMol+STGM} \\ \midrule
albuterol\_similarity     & .885 & .824 & .964 & .940 & .950 & .547 & .501 & .949 & .950 & .947 & .793 & .840 \\
amlodipine\_mpo           & .679 & .680 & .720 & .680 & .733 & .517 & .520 & .713 & .690 & .670 & .580 & .657 \\
celecoxib\_rediscovery    & .664 & .716 & .860 & .820 & .798 & .392 & .426 & .890 & .750 & .856 & .549 & .644 \\
deco\_hop                 & .619 & .710 & .672 & .660 & .748 & .604 & .595 & .914 & .790 & .831 & .780 & .877 \\
drd2                       & .928 & .956 & .902 & .990 & .985 & .634 & .666 & .967 & .970 & .981 & .979 & .993 \\
fexofenadine\_mpo         & .772 & .798 & .806 & .780 & .845 & .713 & .602 & .804 & .860 & .833 & .744 & .796 \\
gsk3b                      & .876 & .896 & .877 & .920 & .952 & .604 & .428 & .944 & .930 & .924 & .905 & .921 \\
isomers\_c7h8n2o2         & .939 & .932 & .911 & .890 & .968 & .927 & .804 & .928 & .970 & .975 & .959 & .950 \\
isomers\_c9h10n2o2pf2cl   & .819 & .864 & .828 & .790 & .874 & .649 & .449 & .874 & .900 & .875 & .809 & .812 \\
jnk3                       & .731 & .806 & .785 & .870 & .825 & .055 & .137 & .818 & .650 & .910 & .797 & .840 \\
median1                    & .291 & .340 & .415 & .350 & .342 & .276 & .131 & .374 & .340 & .357 & .285 & .349 \\
median2                    & .280 & .255 & .408 & .290 & .288 & .230 & .184 & .307 & .270 & .349 & .262 & .343 \\
mestranol\_similarity     & .614 & .658 & .930 & .890 & .797 & .494 & .358 & .780 & .840 & .759 & .451 & .886 \\
osimertinib\_mpo          & .820 & .819 & .833 & .840 & .870 & .780 & .782 & .870 & .870 & .856 & .827 & .841 \\
perindopril\_mpo          & .557 & .584 & .651 & .530 & .645 & .448 & .424 & .669 & .620 & .774 & .513 & .569 \\
qed                        & .936 & .940 & .947 & .940 & .942 & .937 & .802 & .945 & .940 & .940 & .942 & .942 \\
ranolazine\_mpo           & .774 & .812 & .810 & .750 & .848 & .705 & .545 & .780 & .840 & .839 & .754 & .769 \\
scaffold\_hop             & .547 & .531 & .529 & .840 & .589 & .508 & .459 & .790 & .550 & .541 & .561 & .559 \\
sitagliptin\_mpo          & .567 & .501 & .474 & .390 & .709 & .222 & .176 & .673 & .780 & .454 & .451 & .433 \\
thiothixene\_rediscovery  & .514 & .504 & .727 & .580 & .625 & .412 & .336 & .638 & .480 & .647 & .503 & .592 \\
troglitazone\_rediscovery & .380 & .437 & .756 & .520 & .595 & .295 & .292 & .496 & .540 & .579 & .304 & .540 \\
valsartan\_smarts         & .000 & .158 & .000 & .030 & .210 & .000 & .000 & .000 & .000 & .000 & .000 & .000 \\
zaleplon\_mpo             & .515 & .504 & .499 & .520 & .536 & .459 & .403 & .574 & .560 & .529 & .507 & .524 \\ \midrule
sum                        & 14.708 & 15.225 & 16.303 & 15.800 & 16.676 & 11.410 & 10.019 & 16.700 & 16.100 & 16.425 & 14.255 & 15.664 \\ \bottomrule
\end{tabular}}}
\caption{\rev{Additional task-agnostic PMO-10K results. Values shown are source-reported task means except the GP-MoLFormer sum, derived from rounded task means, and the SEISMO no-description sum, reconstructed from released traces. LICO used five development tasks outside the measured curves; ACEGEN may use a different MolScore implementation. SEGO makes 100 proposals and SEISMO makes 50, after which their terminal Top-10 values are held fixed through the 10,000-call horizon. InVirtuoGen denotes its budget-comparable cold-start configuration. Unreported Valsartan AUC values are conservatively assigned zero. ChemTSv3$^{\dagger}$ uses a newer PyTDC release that changes at least the isomer, Sitagliptin, and Zaleplon oracle implementations. GenMol+STGM is a single-seed result at 10,000 calls; its source reports a nonzero terminal Valsartan Top-10 score separately but not the corresponding AUC trajectory, so the zero shown here is a lower bound. Source results may also reflect different PMO oracle-package versions.}}
\label{tab:PMO_full_recent_agnostic}
\endgroup
\end{table}

\begin{table}[!h]
\begingroup
\centering
\scriptsize
\resizebox{\columnwidth}{!}{
\begin{tabular}{lccc>{\color{black}}ccccc}
\toprule
Oracle & \multicolumn{3}{c}{MOLLEO} & ExLLM & SEISMO & \multicolumn{3}{c}{Mol-E} \\
 & MolSTM & BioT5 & GPT-4 & & reconstructed & Chemlactica-125M & Chemlactica-1.3B & Chemma-2B \\ \midrule
albuterol\_similarity     & .929 & .968 & .985 & .989 & 1.000 & .995 $\pm$ .000 & .995 $\pm$ .000 & .995 $\pm$ .000 \\
amlodipine\_mpo           & .674 & .776 & .773 & .874 & .900 & .951 $\pm$ .000 & .951 $\pm$ .000 & .950 $\pm$ .000 \\
celecoxib\_rediscovery    & .594 & .508 & .864 & .891 & .885 & .984 $\pm$ .000 & .984 $\pm$ .000 & .984 $\pm$ .000 \\
deco\_hop                 & .613 & .827 & .942 & .956 & .958 & .986 $\pm$ .002 & .980 $\pm$ .001 & .982 $\pm$ .006 \\
drd2                       & .975 & .981 & .968 & .980 & 1.000 & .979 $\pm$ .003 & .976 $\pm$ .005 & .975 $\pm$ .007 \\
fexofenadine\_mpo         & .789 & .773 & .847 & .984 & .984 & .995 $\pm$ .000 & .995 $\pm$ .000 & .995 $\pm$ .000 \\
gsk3b                      & .898 & .889 & .863 & .818 & .976 & .939 $\pm$ .016 & .947 $\pm$ .014 & .942 $\pm$ .016 \\
isomers\_c7h8n2o2         & .948 & .928 & .984 & .984 & 1.000 & .957 $\pm$ .007 & .851 $\pm$ .083 & .961 $\pm$ .012 \\
isomers\_c9h10n2o2pf2cl   & .871 & .873 & .874 & .961 & 1.000 & .920 $\pm$ .022 & .927 $\pm$ .024 & .946 $\pm$ .017 \\
jnk3                       & .643 & .728 & .790 & .732 & .468 & .877 $\pm$ .047 & .854 $\pm$ .114 & .912 $\pm$ .023 \\
median1                    & .298 & .338 & .352 & .384 & .344 & .465 $\pm$ .035 & .489 $\pm$ .000 & .482 $\pm$ .002 \\
median2                    & .251 & .259 & .275 & .475 & .365 & .483 $\pm$ .005 & .488 $\pm$ .002 & .483 $\pm$ .003 \\
mestranol\_similarity     & .596 & .717 & .972 & .980 & 1.000 & .982 $\pm$ .003 & .995 $\pm$ .000 & .980 $\pm$ .004 \\
osimertinib\_mpo          & .823 & .817 & .835 & .902 & .932 & .992 $\pm$ .001 & .995 $\pm$ .000 & .995 $\pm$ .000 \\
perindopril\_mpo          & .554 & .738 & .600 & .797 & .810 & .918 $\pm$ .002 & .908 $\pm$ .011 & .911 $\pm$ .007 \\
qed                        & .937 & .937 & .948 & .943 & .938 & .943 $\pm$ .000 & .943 $\pm$ .000 & .943 $\pm$ .000 \\
ranolazine\_mpo           & .725 & .749 & .769 & .855 & .935 & .991 $\pm$ .002 & .989 $\pm$ .003 & .990 $\pm$ .002 \\
scaffold\_hop             & .527 & .559 & .971 & .916 & .934 & .989 $\pm$ .001 & .986 $\pm$ .003 & .983 $\pm$ .004 \\
sitagliptin\_mpo          & .548 & .506 & .584 & .555 & .238 & .720 $\pm$ .008 & .704 $\pm$ .044 & .739 $\pm$ .023 \\
thiothixene\_rediscovery  & .508 & .696 & .727 & .910 & .919 & .984 $\pm$ .000 & .985 $\pm$ .000 & .985 $\pm$ .000 \\
troglitazone\_rediscovery & .381 & .390 & .562 & .726 & .815 & .985 $\pm$ .000 & .985 $\pm$ .000 & .985 $\pm$ .000 \\
valsartan\_smarts         & .000 & .000 & .867 & .831 & .141 & .881 $\pm$ .048 & .893 $\pm$ .020 & .845 $\pm$ .060 \\
zaleplon\_mpo             & .475 & .465 & .510 & .723 & .670 & .515 $\pm$ .046 & .372 $\pm$ .090 & .587 $\pm$ .020 \\ \midrule
sum                        & 14.557 & 15.424 & 17.862 & 19.165 & 18.214 & 20.429 $\pm$ .079 & 20.192 $\pm$ .264 & \textbf{20.551 $\pm$ .038} \\ \bottomrule
\end{tabular}}
\caption{Task-informed comparison on all PMO tasks. Mol-E values are mean Top-10 AUC $\uparrow$ $\pm$ population standard deviation over five independent runs; other methods are source-reported means. \rev{ExLLM uses semantic objective descriptions and iterative textual feedback. SEISMO makes 50 proposals; its displayed AUC is reconstructed by flat-filling the released terminal Top-10 values through 10,000 calls.}}
\label{tab:PMO_full_task_informed}
\endgroup
\end{table}

\begin{table}[!h]
\begingroup
\centering
\scriptsize
\resizebox{\columnwidth}{!}{
{\color{gray}
\begin{tabular}{lccccc}
\toprule
Oracle & f-RAG$^{*}$ & GenMol$^{*}$ & \multicolumn{2}{c}{Graph-GRPO$^{*}$} & InVirtuoGen$^{*}$ \\
 & & & no prescreen & prescreened & prescreened \\ \midrule
albuterol\_similarity     & 0.977 $\pm$ 0.002 & 0.937 $\pm$ 0.010 & 0.994 $\pm$ 0.000 & 0.994 $\pm$ 0.000 & .975 \\
amlodipine\_mpo           & 0.749 $\pm$ 0.019 & 0.810 $\pm$ 0.012 & 0.823 $\pm$ 0.008 & 0.823 $\pm$ 0.008 & .836 \\
celecoxib\_rediscovery    & 0.778 $\pm$ 0.007 & 0.826 $\pm$ 0.018 & 0.890 $\pm$ 0.000 & 0.890 $\pm$ 0.000 & .839 \\
deco\_hop                 & 0.936 $\pm$ 0.011 & 0.960 $\pm$ 0.010 & 0.762 $\pm$ 0.127 & 0.942 $\pm$ 0.005 & .968 \\
drd2                       & 0.992 $\pm$ 0.000 & 0.995 $\pm$ 0.000 & 0.995 $\pm$ 0.000 & 0.995 $\pm$ 0.000 & .995 \\
fexofenadine\_mpo         & 0.856 $\pm$ 0.016 & 0.894 $\pm$ 0.028 & 0.984 $\pm$ 0.001 & 0.984 $\pm$ 0.001 & .904 \\
gsk3b                      & 0.969 $\pm$ 0.003 & 0.986 $\pm$ 0.003 & 0.965 $\pm$ 0.006 & 0.948 $\pm$ 0.015 & .988 \\
isomers\_c7h8n2o2         & 0.955 $\pm$ 0.008 & 0.942 $\pm$ 0.004 & 0.995 $\pm$ 0.000 & 0.995 $\pm$ 0.000 & .988 \\
isomers\_c9h10n2o2pf2cl   & 0.850 $\pm$ 0.005 & 0.833 $\pm$ 0.014 & 0.932 $\pm$ 0.011 & 0.932 $\pm$ 0.011 & .898 \\
jnk3                       & 0.904 $\pm$ 0.004 & 0.906 $\pm$ 0.023 & 0.910 $\pm$ 0.019 & 0.910 $\pm$ 0.019 & .898 \\
median1                    & 0.340 $\pm$ 0.007 & 0.398 $\pm$ 0.000 & 0.388 $\pm$ 0.000 & 0.388 $\pm$ 0.000 & .386 \\
median2                    & 0.323 $\pm$ 0.005 & 0.359 $\pm$ 0.004 & 0.300 $\pm$ 0.002 & 0.322 $\pm$ 0.039 & .377 \\
mestranol\_similarity     & 0.671 $\pm$ 0.021 & 0.982 $\pm$ 0.000 & 0.974 $\pm$ 0.002 & 0.980 $\pm$ 0.002 & .991 \\
osimertinib\_mpo          & 0.866 $\pm$ 0.009 & 0.876 $\pm$ 0.008 & 0.922 $\pm$ 0.002 & 0.924 $\pm$ 0.003 & .881 \\
perindopril\_mpo          & 0.681 $\pm$ 0.017 & 0.718 $\pm$ 0.012 & 0.689 $\pm$ 0.036 & 0.690 $\pm$ 0.033 & .753 \\
qed                        & 0.939 $\pm$ 0.001 & 0.942 $\pm$ 0.000 & 0.944 $\pm$ 0.000 & 0.944 $\pm$ 0.000 & .943 \\
ranolazine\_mpo           & 0.820 $\pm$ 0.016 & 0.821 $\pm$ 0.011 & 0.928 $\pm$ 0.001 & 0.928 $\pm$ 0.001 & .854 \\
scaffold\_hop             & 0.576 $\pm$ 0.014 & 0.628 $\pm$ 0.008 & 0.622 $\pm$ 0.015 & 0.711 $\pm$ 0.113 & .711 \\
sitagliptin\_mpo          & 0.601 $\pm$ 0.011 & 0.584 $\pm$ 0.034 & 0.879 $\pm$ 0.014 & 0.879 $\pm$ 0.014 & .743 \\
thiothixene\_rediscovery  & 0.584 $\pm$ 0.009 & 0.692 $\pm$ 0.123 & 0.842 $\pm$ 0.001 & 0.842 $\pm$ 0.001 & .652 \\
troglitazone\_rediscovery & 0.448 $\pm$ 0.017 & 0.867 $\pm$ 0.022 & 0.711 $\pm$ 0.008 & 0.711 $\pm$ 0.008 & .853 \\
valsartan\_smarts         & 0.627 $\pm$ 0.058 & 0.822 $\pm$ 0.042 & 0.841 $\pm$ 0.019 & 0.841 $\pm$ 0.019 & .935 \\
zaleplon\_mpo             & 0.486 $\pm$ 0.004 & 0.584 $\pm$ 0.011 & 0.697 $\pm$ 0.004 & 0.697 $\pm$ 0.004 & .624 \\ \midrule
sum                        & 16.928 & 18.362 & 18.987 & 19.270 & 18.993 \\ \bottomrule
\end{tabular}}}
\caption{\rev{PMO-10K results with uncounted target-oracle use.} Gray results are source-reported means and are not directly comparable under PMO oracle-budget accounting. f-RAG and GenMol omit approximately 250,000 target-oracle prescreening calls per task; Graph-GRPO omits target-oracle policy-training calls, and its prescreened variant additionally uses 250,000 prescreening calls. \rev{The InVirtuoGen prescreened variant likewise omits approximately 250,000 target-oracle calls; only its cold-start configuration is included in the comparable tables.}}
\label{tab:PMO_full_budget_noncomparable}
\endgroup
\end{table}

\clearpage
\subsubsection{PMO at a 1,000-call horizon}

\rev{A 1,000-call horizon has become increasingly common as a more stringent test of sample efficiency. Table \ref{tab:PMO_1k_external} compares recent external results, while Table \ref{tab:PMO_1k_mole} recomputes all six Mol-E configurations from their saved trajectories using the PMO trapezoidal AUC rule at calls 0, 100, \ldots, 1,000. These values are post hoc truncations of runs whose hyperparameters were selected for PMO-10K rather than retuned for the 1,000-call setting. Moreover, inspection of the saved trajectories shows that most runs do not trigger online language-model fine-tuning while generating their first 1,000 molecules, so the PMO-1K results often precede any oracle-dependent update of the model weights. Task-informed SEISMO obtains the highest comparable full-23-task sum, 18.156. Mol-E with Chemlactica-1.3B in the task-informed regime scores 16.469 and is second among methods covering all 23 tasks without uncounted target-oracle prescreening. Partial-task and prescreened results are retained with explicit qualifications rather than included in this ranking.}

\begin{table}[!h]
\begingroup
\centering
\scriptsize
\resizebox{\columnwidth}{!}{
{\color{black}
\begin{tabular}{lcccccccc}
\toprule
Oracle & LICO & ExLLM & MT-Mol$^{*}$ & MolLIBRA-L & FORGE & INDIBATOR & SEGO & SEISMO \\ \midrule
albuterol\_similarity     & .656 & .886 & .998 & .974 & .779 & -- & .537 & .999 \\
amlodipine\_mpo           & .541 & .784 & .647 & .619 & .551 & .845 & .509 & .898 \\
celecoxib\_rediscovery    & .447 & .634 & .867 & .709 & .428 & .821 & .385 & .883 \\
deco\_hop                 & .596 & .855 & .842 & .633 & .624 & -- & .602 & .956 \\
drd2                       & .859 & .797 & .756 & .967 & .816 & .950 & .605 & .999 \\
fexofenadine\_mpo         & .700 & .893 & .883 & .741 & .740 & .925 & .703 & .983 \\
gsk3b                      & .617 & .449 & .308 & .723 & .620 & .942 & .577 & .974 \\
isomers\_c7h8n2o2         & .779 & .842 & .986 & .841 & .723 & -- & .896 & .999 \\
isomers\_c9h10n2o2pf2cl   & .672 & .746 & .914 & .828 & .713 & -- & .624 & .998 \\
jnk3                       & .336 & .308 & .125 & .495 & .611 & .914 & .054 & .464 \\
median1                    & .217 & .329 & .321 & .263 & .271 & -- & .269 & .343 \\
median2                    & .193 & .370 & .322 & .282 & .221 & -- & .226 & .365 \\
mestranol\_similarity     & .423 & .798 & .996 & .630 & .450 & -- & .484 & .999 \\
osimertinib\_mpo          & .759 & .789 & .796 & .828 & .794 & .941 & .769 & .924 \\
perindopril\_mpo          & .473 & .660 & .542 & .526 & .509 & .775 & .440 & .809 \\
qed                        & .925 & .896 & .903 & .932 & .937 & -- & .932 & .937 \\
ranolazine\_mpo           & .687 & .662 & .233 & .777 & .680 & .848 & .690 & .926 \\
scaffold\_hop             & .480 & .786 & .646 & .536 & .532 & -- & .505 & .931 \\
sitagliptin\_mpo          & .315 & .290 & .067 & .434 & .300 & .225 & .215 & .234 \\
thiothixene\_rediscovery  & .343 & .718 & .719 & .615 & .400 & .831 & .403 & .919 \\
troglitazone\_rediscovery & .292 & .348 & .841 & .412 & .288 & .838 & .289 & .813 \\
valsartan\_smarts         & .000 & .069 & .000 & -- & .000 & -- & .000 & .137 \\
zaleplon\_mpo             & .404 & .622 & .625 & .445 & .435 & .730 & .447 & .666 \\ \midrule
sum                        & 11.714 & 14.533 & 15.420 & 14.208 (22) & 12.420 & 10.585 (13) & 11.161 & \textbf{18.156} \\ \bottomrule
\end{tabular}}}
\caption{\rev{External PMO-1K Top-10 AUC results. SEISMO receives semantic task descriptions and explanatory feedback and makes 50 proposals; SEGO makes 100 proposals and reports its flat-filled AUC. MolLIBRA omits Valsartan, and INDIBATOR reports 13 tasks. FORGE describes 22 tasks but displays Valsartan as zero. $^{*}$MT-Mol retrieves oracle-scored prompt examples from ZINC250K, omitting roughly 250,000 target-oracle prescreening calls per task from its nominal 1,000-call budget.}}
\label{tab:PMO_1k_external}
\endgroup
\end{table}

\begin{table}[!h]
\begingroup
\centering
\scriptsize
\resizebox{\columnwidth}{!}{
{\color{black}
\begin{tabular}{lccc|ccc}
\toprule
 & \multicolumn{3}{c}{Task-agnostic Mol-E} & \multicolumn{3}{c}{Task-informed Mol-E} \\
Oracle & Chemlactica-125M & Chemlactica-1.3B & Chemma-2B & Chemlactica-125M & Chemlactica-1.3B & Chemma-2B \\ \midrule
albuterol\_similarity     & .700 $\pm$ .039 & .553 $\pm$ .051 & .614 $\pm$ .036 & .950 $\pm$ .000 & .950 $\pm$ .000 & .950 $\pm$ .000 \\
amlodipine\_mpo           & .513 $\pm$ .026 & .495 $\pm$ .016 & .505 $\pm$ .024 & .891 $\pm$ .003 & .895 $\pm$ .001 & .888 $\pm$ .002 \\
celecoxib\_rediscovery    & .432 $\pm$ .034 & .380 $\pm$ .029 & .425 $\pm$ .034 & .939 $\pm$ .000 & .939 $\pm$ .000 & .938 $\pm$ .000 \\
deco\_hop                 & .585 $\pm$ .016 & .621 $\pm$ .057 & .573 $\pm$ .005 & .869 $\pm$ .016 & .845 $\pm$ .009 & .873 $\pm$ .010 \\
drd2                       & .789 $\pm$ .031 & .762 $\pm$ .046 & .750 $\pm$ .066 & .789 $\pm$ .031 & .762 $\pm$ .046 & .750 $\pm$ .066 \\
fexofenadine\_mpo         & .669 $\pm$ .018 & .672 $\pm$ .013 & .671 $\pm$ .009 & .949 $\pm$ .001 & .949 $\pm$ .000 & .950 $\pm$ .000 \\
gsk3b                      & .596 $\pm$ .073 & .618 $\pm$ .043 & .634 $\pm$ .065 & .596 $\pm$ .073 & .618 $\pm$ .043 & .634 $\pm$ .065 \\
isomers\_c7h8n2o2         & .284 $\pm$ .110 & .351 $\pm$ .086 & .333 $\pm$ .081 & .625 $\pm$ .044 & .616 $\pm$ .020 & .686 $\pm$ .068 \\
isomers\_c9h10n2o2pf2cl   & .305 $\pm$ .174 & .160 $\pm$ .078 & .292 $\pm$ .061 & .473 $\pm$ .023 & .577 $\pm$ .076 & .601 $\pm$ .111 \\
jnk3                       & .408 $\pm$ .106 & .444 $\pm$ .069 & .453 $\pm$ .074 & .408 $\pm$ .106 & .444 $\pm$ .069 & .446 $\pm$ .080 \\
median1                    & .216 $\pm$ .020 & .200 $\pm$ .017 & .209 $\pm$ .014 & .399 $\pm$ .019 & .454 $\pm$ .001 & .396 $\pm$ .010 \\
median2                    & .215 $\pm$ .015 & .207 $\pm$ .009 & .203 $\pm$ .019 & .403 $\pm$ .007 & .440 $\pm$ .002 & .377 $\pm$ .011 \\
mestranol\_similarity     & .536 $\pm$ .117 & .460 $\pm$ .062 & .405 $\pm$ .028 & .821 $\pm$ .028 & .950 $\pm$ .000 & .800 $\pm$ .040 \\
osimertinib\_mpo          & .731 $\pm$ .005 & .739 $\pm$ .005 & .738 $\pm$ .009 & .924 $\pm$ .010 & .949 $\pm$ .001 & .950 $\pm$ .000 \\
perindopril\_mpo          & .441 $\pm$ .012 & .441 $\pm$ .013 & .458 $\pm$ .013 & .821 $\pm$ .004 & .802 $\pm$ .008 & .802 $\pm$ .004 \\
qed                        & .888 $\pm$ .002 & .886 $\pm$ .003 & .888 $\pm$ .002 & .899 $\pm$ .001 & .900 $\pm$ .001 & .900 $\pm$ .000 \\
ranolazine\_mpo           & .662 $\pm$ .023 & .690 $\pm$ .016 & .682 $\pm$ .013 & .908 $\pm$ .016 & .891 $\pm$ .027 & .899 $\pm$ .021 \\
scaffold\_hop             & .479 $\pm$ .022 & .481 $\pm$ .011 & .467 $\pm$ .009 & .885 $\pm$ .010 & .859 $\pm$ .033 & .830 $\pm$ .040 \\
sitagliptin\_mpo          & .083 $\pm$ .079 & .032 $\pm$ .010 & .094 $\pm$ .044 & .338 $\pm$ .040 & .363 $\pm$ .037 & .379 $\pm$ .029 \\
thiothixene\_rediscovery  & .367 $\pm$ .020 & .332 $\pm$ .024 & .358 $\pm$ .021 & .939 $\pm$ .000 & .939 $\pm$ .000 & .940 $\pm$ .000 \\
troglitazone\_rediscovery & .296 $\pm$ .019 & .274 $\pm$ .024 & .273 $\pm$ .010 & .940 $\pm$ .000 & .940 $\pm$ .000 & .940 $\pm$ .000 \\
valsartan\_smarts         & .000 $\pm$ .000 & .000 $\pm$ .000 & .017 $\pm$ .033 & .000 $\pm$ .000 & .382 $\pm$ .023 & .181 $\pm$ .070 \\
zaleplon\_mpo             & .188 $\pm$ .040 & .118 $\pm$ .055 & .170 $\pm$ .022 & .087 $\pm$ .057 & .006 $\pm$ .007 & .091 $\pm$ .049 \\ \midrule
sum                        & 10.381 $\pm$ .319 & 9.919 $\pm$ .173 & 10.213 $\pm$ .166 & 15.854 $\pm$ .224 & \textbf{16.469 $\pm$ .158} & 16.202 $\pm$ .204 \\ \bottomrule
\end{tabular}}}
\caption{\rev{Mol-E PMO-1K results recomputed from saved trajectories. Entries are mean Top-10 AUC $\uparrow$ $\pm$ population standard deviation over five seeds; total deviations are computed from seed-wise sums. The same optimization runs underlie the 10K results. At this horizon, only the task-informed Chemma-2B QED seed 41 ends before 1,000 unique calls, and its terminal Top-10 value is flat-filled from call 800 under the benchmark convention.}}
\label{tab:PMO_1k_mole}
\endgroup
\end{table}

\clearpage

\subsection{Analysis of Molecular Optimization}
\subsubsection{\rev{Ablation of Online Fine-Tuning}}
\label{sec:no-ft-ablation}
\begin{figure}
    \centering\includegraphics[width=1.0\linewidth]{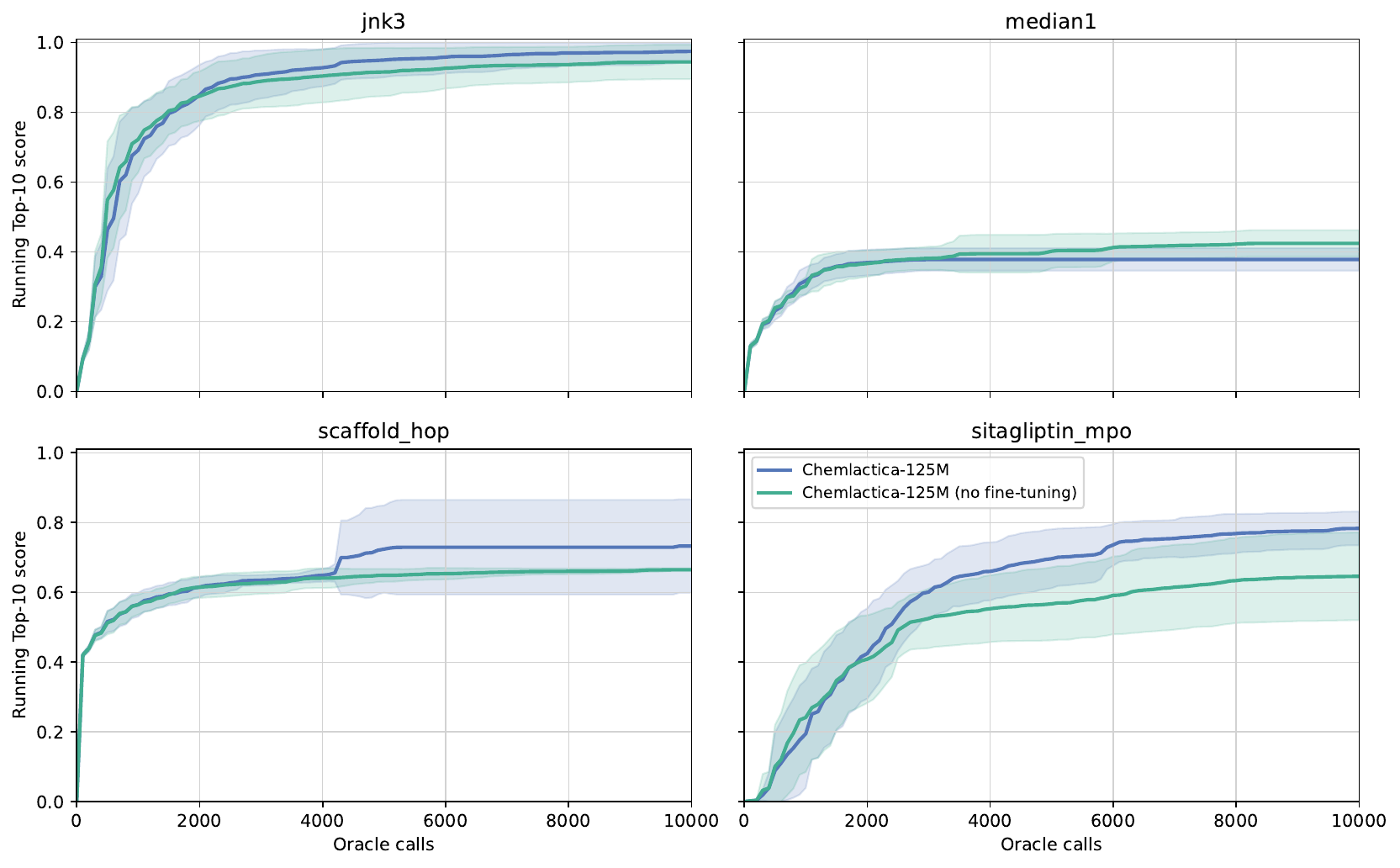}
    \caption{\rev{Running Top-10 oracle score for Chemlactica-125M with and without online fine-tuning. Curves show the mean $\pm$ population standard deviation over five seeds.}}
    \label{fig:no-ft-chemlactica-125m}
\end{figure}

\begin{figure}
    \centering\includegraphics[width=1.0\linewidth]{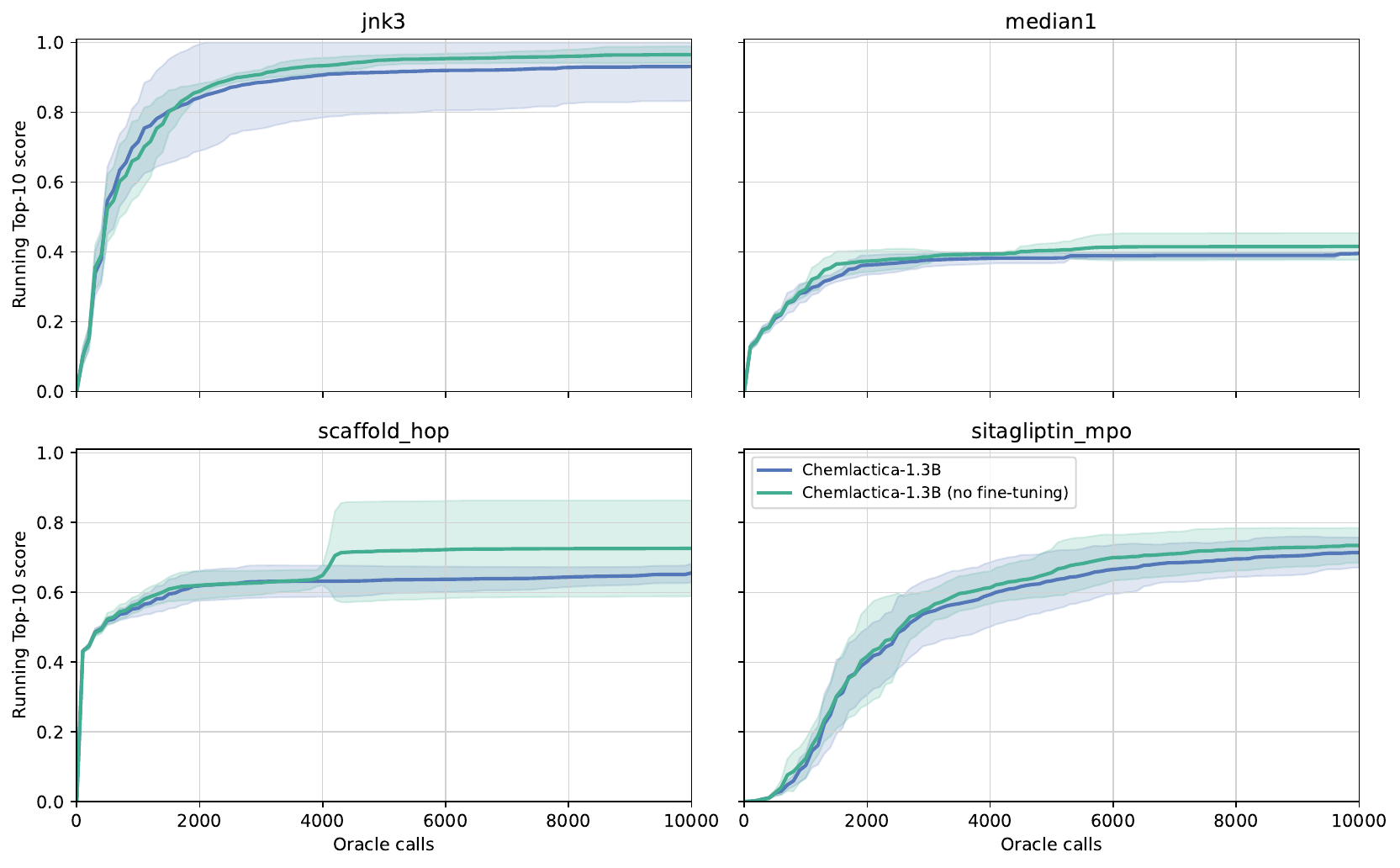}
    \caption{\rev{Running Top-10 oracle score for Chemlactica-1.3B with and without online fine-tuning. Curves show the mean $\pm$ population standard deviation over five seeds.}}
    \label{fig:no-ft-chemlactica-1.3b}
\end{figure}

\begin{figure}
    \centering\includegraphics[width=1.0\linewidth]{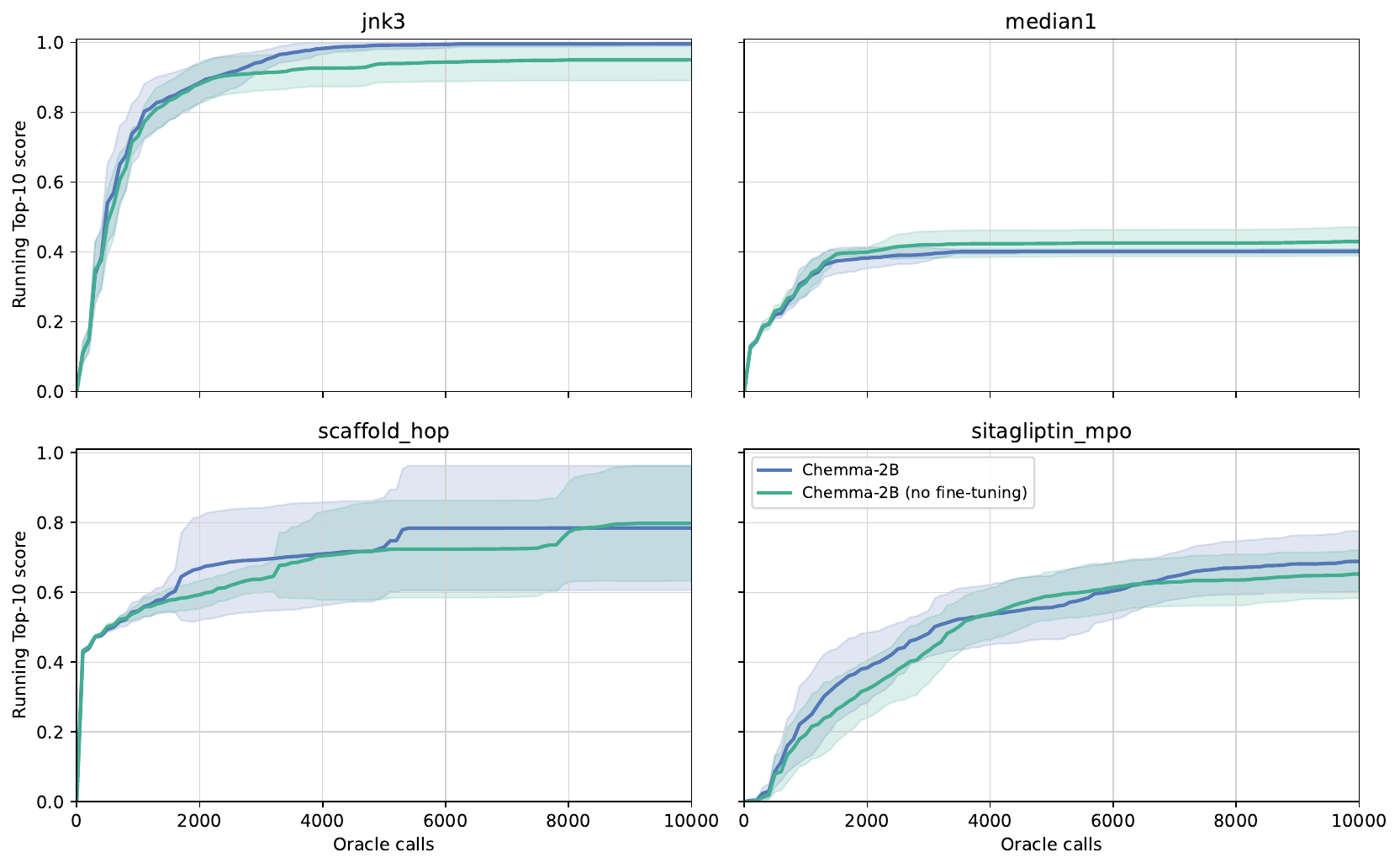}
    \caption{\rev{Running Top-10 oracle score for Chemma-2B with and without online fine-tuning. Curves show the mean $\pm$ population standard deviation over five seeds.}}
    \label{fig:no-ft-chemma-2b}
\end{figure}

\rev{To isolate the effect of online fine-tuning, we reran four PMO tasks (\texttt{jnk3}, \texttt{median1}, \texttt{scaffold\_hop}, and \texttt{sitagliptin\_mpo}) with online fine-tuning disabled. We retained the five reproduction seeds, the corrected Sitagliptin oracle, and all other model-specific generation and search settings. Table \ref{tab:abl_optim_algo} reports the resulting AUC values, while Figures \ref{fig:no-ft-chemlactica-125m}--\ref{fig:no-ft-chemma-2b} show the corresponding running Top-10 trajectories.}

\rev{The effect is task- and backbone-dependent. Online fine-tuning increases the four-task sum for Chemlactica-125M (2.507 versus 2.377) and Chemma-2B (2.515 versus 2.448), but decreases it for Chemlactica-1.3B (2.372 versus 2.484). The task-level comparisons are likewise mixed. Thus, online fine-tuning can improve Mol-E, particularly for the corrected Sitagliptin task with Chemlactica-125M, but it does not provide a uniform benefit across models and tasks.}
\begin{center}
\resizebox{\columnwidth}{!}{{\color{black}\begin{tabular}{ccc|cc|cc}
\toprule
             & \multicolumn{2}{c}{Chemlactica-125M}  & \multicolumn{2}{|c|}{Chemlactica-1.3B}  & \multicolumn{2}{c}{Chemma-2B}         \\
             & fine-tuning       & disabled    & fine-tuning       & disabled    & fine-tuning       & disabled    \\ \midrule
jnk3             & 0.877 $\pm$ 0.047 & 0.859 $\pm$ 0.055 & 0.854 $\pm$ 0.114 & 0.877 $\pm$ 0.014 & 0.914 $\pm$ 0.023 & 0.875 $\pm$ 0.044 \\
median1          & 0.359 $\pm$ 0.030 & 0.382 $\pm$ 0.034 & 0.361 $\pm$ 0.009 & 0.379 $\pm$ 0.022 & 0.377 $\pm$ 0.009 & 0.398 $\pm$ 0.033 \\
scaffold\_hop    & 0.668 $\pm$ 0.082 & 0.626 $\pm$ 0.017 & 0.617 $\pm$ 0.032 & 0.667 $\pm$ 0.089 & 0.709 $\pm$ 0.125 & 0.682 $\pm$ 0.093 \\
sitagliptin\_mpo & 0.603 $\pm$ 0.048 & 0.510 $\pm$ 0.094 & 0.539 $\pm$ 0.062 & 0.561 $\pm$ 0.053 & 0.515 $\pm$ 0.062 & 0.493 $\pm$ 0.047 \\ \midrule
sum              & \textbf{2.507 $\pm$ 0.089} & 2.377 $\pm$ 0.114 & 2.372 $\pm$ 0.140 & \textbf{2.484 $\pm$ 0.140} & \textbf{2.515 $\pm$ 0.180} & 2.448 $\pm$ 0.098 \\ \bottomrule
\end{tabular}}}
\captionof{table}{\rev{Online fine-tuning ablation on four PMO tasks. Entries are mean Top-10 AUC $\uparrow$ $\pm$ population standard deviation over five seeds; total deviations are computed from seed-wise sums. ``Disabled'' changes only the online fine-tuning step.}}
\label{tab:abl_optim_algo}
\end{center}

\subsection{Task-Informed PMO Prompt Registry}
\label{sec:pmo-task-informed-prompts}

Table \ref{tab:pmo-task-informed-prefixes} lists the fixed prefix used for every task-informed run. Each prefix uses only tags present in the language model's pretraining data. The registry was fixed before evaluation and was shared across Chemlactica-125M, Chemlactica-1.3B, and Chemma-2B.

\begingroup
\setlength{\tabcolsep}{3pt}
\renewcommand{\arraystretch}{0.5}
\begin{longtable}{@{}>{\fontsize{4.8}{4.8}\selectfont\raggedright\arraybackslash}p{0.16\textwidth}>{\fontsize{4.8}{4.8}\selectfont\raggedright\arraybackslash}p{0.80\textwidth}@{}}
\caption{Exact fixed prefixes used in the task-informed PMO experiments. An empty entry means that the task-informed and task-agnostic generation prompts are identical.}
\label{tab:pmo-task-informed-prefixes}\\
\toprule
PMO task & Prompt prefix \\
\midrule
\endfirsthead
\multicolumn{2}{c}{\tablename\ \thetable\ (continued)}\\
\toprule
PMO task & Prompt prefix \\
\midrule
\endhead
\midrule
\multicolumn{2}{r}{Continued on next page}\\
\endfoot
\bottomrule
\endlastfoot
\texttt{albuterol\_similarity} & \nolinkurl{[SIMILAR]CC(C)(C)NCC(O)c1ccc(O)c(CO)c1 0.75[/SIMILAR]} \\
\texttt{amlodipine\_mpo} & \nolinkurl{[SIMILAR]CCOC(=O)C1=C(COCCN)NC(C)=C(C(=O)OC)C1c1ccccc1Cl 0.99[/SIMILAR][RINGCOUNT]3[/RINGCOUNT]} \\
\texttt{celecoxib\_rediscovery} & \nolinkurl{[SIMILAR]Cc1ccc(-c2cc(C(F)(F)F)nn2-c2ccc(S(N)(=O)=O)cc2)cc1 0.99[/SIMILAR]} \\
\texttt{deco\_hop} & \nolinkurl{[SIMILAR]CCCOc1cc2ncnc(Nc3ccc4ncsc4c3)c2cc1S(=O)(=O)C(C)(C)C 0.85[/SIMILAR]} \\
\texttt{drd2} & \emph{Empty} \\
\texttt{fexofenadine\_mpo} & \nolinkurl{[SIMILAR]CC(C)(C(=O)O)c1ccc(C(O)CCCN2CCC(C(O)(c3ccccc3)c3ccccc3)CC2)cc1 0.80[/SIMILAR][TPSA]90.00[/TPSA][CLOGP]4.00[/CLOGP]} \\
\texttt{gsk3b} & \emph{Empty} \\
\texttt{isomers\_c7h8n2o2} & \nolinkurl{[FORMULA]C7H8N2O2[/FORMULA]} \\
\texttt{isomers\_c9h10n2o2pf2cl} & \nolinkurl{[FORMULA]C9H10N2O2PF2Cl[/FORMULA]} \\
\texttt{jnk3} & \emph{Empty} \\
\texttt{median1} & \nolinkurl{[SIMILAR]CC12CCC(CC1=O)C2(C)C 0.55[/SIMILAR][SIMILAR]CC1CCC(C(C)C)C(O)C1 0.55[/SIMILAR]} \\
\texttt{median2} & \nolinkurl{[SIMILAR]CN1CC(=O)N2C(Cc3c([nH]c4ccccc34)C2c2ccc3c(c2)OCO3)C1=O 0.57[/SIMILAR][SIMILAR]CCCc1nn(C)c2c(=O)[nH]c(-c3cc(S(=O)(=O)N4CCN(C)CC4)ccc3OCC)nc12 0.57[/SIMILAR]} \\
\texttt{mestranol\_similarity} & \nolinkurl{[SIMILAR]C#C[C@]1(O)CC[C@H]2[C@@H]3CCc4cc(OC)ccc4[C@H]3CC[C@@]21C 0.75[/SIMILAR]} \\
\texttt{osimertinib\_mpo} & \nolinkurl{[SIMILAR]C=CC(=O)Nc1cc(Nc2nccc(-c3cn(C)c4ccccc34)n2)c(OC)cc1N(C)CCN(C)C 0.80[/SIMILAR][TPSA]100.00[/TPSA][CLOGP]1.00[/CLOGP]} \\
\texttt{perindopril\_mpo} & \nolinkurl{[SIMILAR]CCCC(NC(C)C(=O)N1C(C(=O)O)CC2CCCCC21)C(=O)OCC 0.99[/SIMILAR][NUMAROMATICRINGS]2[/NUMAROMATICRINGS]} \\
\texttt{qed} & \nolinkurl{[QED]0.95[/QED]} \\
\texttt{ranolazine\_mpo} & \nolinkurl{[SIMILAR]COc1ccccc1OCC(O)CN1CCN(CC(=O)Nc2c(C)cccc2C)CC1 0.70[/SIMILAR][TPSA]95.00[/TPSA][CLOGP]7.00[/CLOGP]} \\
\texttt{scaffold\_hop} & \nolinkurl{[SIMILAR]CCCOc1cc2ncnc(Nc3ccc4ncsc4c3)c2cc1S(=O)(=O)C(C)(C)C 0.75[/SIMILAR]} \\
\texttt{sitagliptin\_mpo} & \nolinkurl{[CLOGP]2.02[/CLOGP][TPSA]77.04[/TPSA][FORMULA]C16H15F6N5O[/FORMULA]} \\
\texttt{thiothixene\_rediscovery} & \nolinkurl{[SIMILAR]CN1CCN(CCC=C2c3ccccc3Sc3ccc(S(=O)(=O)N(C)C)cc32)CC1 0.99[/SIMILAR]} \\
\texttt{troglitazone\_rediscovery} & \nolinkurl{[SIMILAR]Cc1c(C)c2c(c(C)c1O)CCC(C)(COc1ccc(CC3SC(=O)NC3=O)cc1)O2 0.99[/SIMILAR]} \\
\texttt{valsartan\_smarts} & \nolinkurl{[SIMILAR]CN(C=O)Cc1ccc(c2ccccc2)cc1 0.99[/SIMILAR][CLOGP]2.02[/CLOGP][TPSA]77.04[/TPSA]} \\
\texttt{zaleplon\_mpo} & \nolinkurl{[SIMILAR]CCN(C(C)=O)c1cccc(-c2ccnc3c(C#N)cnn23)c1 0.99[/SIMILAR][FORMULA]C19H17N3O2[/FORMULA]} \\
\end{longtable}
\endgroup

We leave the prefixes empty for \texttt{jnk3}, \texttt{gsk3b}, and \texttt{drd2}, because the models were not pretrained on tags that identify these activity oracles. Unsupported conditions (arbitrary SMARTS predicates, Bertz complexity, fluorine counts, and inequality syntax) are not expressed as new tags.

Two cases require special handling. For \texttt{sitagliptin\_mpo}, the oracle rewards low similarity to sitagliptin, so we omit a similarity tag rather than condition on a near-zero similarity value (model's performance at near-zero similarity conditioning is not properly validated). For \texttt{valsartan\_smarts}, we place only the molecular fragment from the oracle's SMARTS condition inside the learned \texttt{[SIMILAR]} tag.

\subsection{The Impact of Floating Point Precision on Molecular Optimization}
\label{subsec:precision}

\subsubsection{Numerical Precision in Model Training}
Lower precision training, including mixed and half-precision methods, is commonly used to increase training throughput. These techniques, employed during our models' pretraining stages, typically have negligible impact on performance and may even provide a regularizing effect\citep{mixedprecision}. However, in molecular optimization involving multiple rounds of fine-tuning, lower numerical precision leads to significantly degraded performance. Several factors contribute to this phenomenon in the specific case of molecular optimization with language models.

\subsubsection{Challenges in Batched Generation}
Molecular optimization pipelines require repeated model calls for generation, followed by oracle function scoring. While batched processing accelerates this process through GPU parallelization, it introduces complications. The necessary padding for batch processing alters matrix sizes, affecting multiply-accumulate operations within the model. These small errors accumulate as they propagate through the model's layers. Lower precision exacerbates these errors, leading to larger discrepancies in logit values and, consequently, more significant impacts on the generated molecules.


\subsubsection{Cascading Effects of Sub-optimal Generations}
In our approach, high-scoring generated molecules are leveraged for fine-tuning and generating similar structures that steer the optimization process. Thus, when lower precision leads to sub-optimal molecule generation, it creates a positive feedback loop. The model is fine-tuned on and guided by these lower-quality molecules, hindering the generation of higher-scoring molecules in subsequent iterations. This causal relationship between successive generations underlies the adverse effects of low-precision training and inference in molecular optimization pipelines.

\subsubsection{Precision Ablation Study}
To quantify the impact of numerical precision on the optimization process, we conducted an ablation study comparing 32-bit floating point precision with bfloat16 precision. Table \ref{tab:low-precision-docking} presents the results of this comparison across all drug discovery case studies described in Section \ref{subsec:docking1}. We show that for the majority of task-metric combinations, optimization results were better when model parameters were in full floating point precision. Despite the potential computational costs, these results demonstrate the importance of maintaining higher numerical precision in molecular optimization tasks.

\begin{table}
\centering
\resizebox{0.6\columnwidth}{!}{
\begin{tabular}{llrr}
\toprule
Metric                                                & Target & Chemlactica-125M        & Chemlactica-125M         \\
                                                      &        & BF16                    & FP32 \\ \midrule
Generative Yield 0.7 $\uparrow$                       & DRD2   & 3501 $\pm$ 252          & \textbf{3733 $\pm$ 512}  \\
                                                      & MK2    & 3000 $\pm$ 80           & \textbf{3772 $\pm$ 578}  \\
                                                      & AChE   & \textbf{4337 $\pm$ 133} & 4108 $\pm$ 67            \\ \midrule
\multirow{3}{*}{Generative Yield 0.8 $\uparrow$}      & DRD2   & 2574 $\pm$ 103          & \textbf{2827 $\pm$ 510}  \\
                                                      & MK2    & 1223 $\pm$ 519          & \textbf{2569 $\pm$ 1156} \\
                                                      & AChE   & 3877 $\pm$ 272          & 3246 $\pm$ 168           \\ \midrule
\multirow{3}{*}{Oracle burden 0.8 (1) $\downarrow$}   & DRD2   & 156 $\pm$ 100           & \textbf{20 $\pm$ 29}     \\
                                                      & MK2    & \textbf{320 $\pm$ 83}   & 345 $\pm$ 312            \\
                                                      & AChE   & \textbf{10 $\pm$ 8}     & 22 $\pm$ 28              \\ \midrule
\multirow{3}{*}{Oracle burden 0.8 (10) $\downarrow$}  & DRD2   & 283 $\pm$ 61            & \textbf{114 $\pm$ 08}    \\
                                                      & MK2    & 631 $\pm$ 100           & \textbf{493 $\pm$ 418}   \\
                                                      & AChE   & \textbf{123 $\pm$ 119}  & 224 $\pm$ 17             \\ \midrule
\multirow{3}{*}{Oracle burden 0.8 (100) $\downarrow$} & DRD2   & 577 $\pm$ 71            & \textbf{364 $\pm$ 119}   \\
                                                      & MK2    & 1134 $\pm$ 178          & \textbf{865 $\pm$ 533}   \\
                                                      & AChE   & \textbf{350 $\pm$ 137}  & 497 $\pm$ 58             \\ \bottomrule
\end{tabular}}
\caption{Impact of numerical precision on Docking MPO experiments from \ref{subsec:docking1}. Oracle burden and generative yield values are reward-threshold dependent. The numbers next to the metrics correspond to the thresholds, and the values in parentheses for oracle burden indicate how many unique molecules need to be generated. The best performance on each task-metric combination is bolded. We use the best-performing hyperparameters from the PMO benchmark.}
\label{tab:low-precision-docking}
\end{table}

\subsection{Property Prediction}
\label{sec:property-predicion-apdx}

\subsubsection{\rev{Supervised fine-tuning recipe.}}  \rev{Inspired by instruction-tuning methodologies and \cite{zhou2023lima}, we generated a specialized training corpus formatted as follows:} \\

\noindent \prompt{[START\_SMILES]$m^{smiles}$[END\_SMILES][PROPERTY]activity {<VALUE>}[/PROPERTY]}.

\subsubsection{\rev{Data splits and hyperparameters.}}
\rev{We use the six official Polaris single-task benchmarks derived from the public Fang et al. ADME data \citep{admet-industrial,cas_wognum_2024_13652588}, retaining the provided preprocessing and train--test partitions. The held-out test sets are used only for final evaluation. For model selection, we create a random validation split containing 20\% of each official training set; after selecting hyperparameters, we retrain on the complete training set. We optimize only the generated response following the \texttt{[PROPERTY]} tag. The search considers learning rates $[10^{-5}, 5\!\times\!10^{-5}, 10^{-4}, 2\!\times\!10^{-4}]$, $[10,15,20]$ epochs, $[0,1,2,3]$ warm-up epochs, and NEFTune noise \citep{jain2023neftune} in $[0,5,10]$. We use a maximum sequence length of 128 and batch size 32, reduced to 16 when required by GPU memory. Table \ref{tab:SFT-hyp} reports the selected settings.}

\begin{table}[b]
\centering
\resizebox{\columnwidth}{!}{%
\begin{tabular}{cllcccccccccccc}
\toprule
\multicolumn{3}{c}{}         & \multicolumn{4}{c}{Chemlactica-125M} & \multicolumn{4}{c}{Chemlactica-1.3B} & \multicolumn{4}{c}{Chemma-2B} \\
\multicolumn{3}{c}{Task}     & LR        & WU    & Ep.    & Nef.    & LR        & WU    & Ep.   & Nef.   & LR       & WU  & Ep.  & Nef.  \\ \midrule
\multicolumn{3}{c}{RLM}      & 5.0e-5    & 0     & 15     & 0       & 1.0e-5    & 3     & 10    & 5      & 2.0e-4   & 1   & 20   & 10    \\
\multicolumn{3}{c}{HLM}      & 5.0e-5    & 1     & 10     & 5       & 1.0e-5    & 3     & 10    & 0      & 1.0e-4   & 3   & 20   & 0     \\
\multicolumn{3}{c}{MDR}      & 5.0e-5    & 2     & 20     & 0       & 1.0e-5    & 1     & 15    & 0      & 2.0e-4   & 1   & 20   & 5     \\
\multicolumn{3}{c}{RPPB}     & 1.0e-4    & 1     & 15     & 0       & 1.0e-5    & 3     & 10    & 0      & 1.0e-4   & 1   & 15   & 5     \\
\multicolumn{3}{c}{HPPB}     & 2.0e-4    & 3     & 20     & 10      & 1.0e-4    & 2     & 15    & 10     & 2.0e-4   & 2   & 15   & 0     \\
\multicolumn{3}{c}{SOL}      & 1.0e-4    & 2     & 10     & 0       & 5.0e-5    & 0     & 20    & 5      & 2.0e-4   & 1   & 15   & 10    \\ \bottomrule
\end{tabular}}
\caption{\rev{Selected hyperparameters for the six Polaris/Fang ADME tasks. LR denotes learning rate, WU warm-up epochs, Ep. the total number of epochs, and Nef. NEFTune noise.}}
\label{tab:SFT-hyp}
\end{table}
\subsection{Model Calibration}
\label{subsec:calibration}

\subsubsection{Methodology}
Model calibration in language modeling refers to the alignment between a model's predicted probabilities for generating specific text and the actual likelihood of that text being correct. To assess the calibration of our models, we developed a suite of multiple-choice property prediction questions based on our training data format. \\

\noindent We generated 2000 questions for each computed property, resulting in 10,000 responses. Each question presented a SMILES string as input: \\

\noindent \prompt{[START\_SMILES]$m^{smiles}$[END\_SMILES]} \\

\noindent and was followed by five potential continuations, with only one being correct. An example of such a continuation for a question testing QED could be:
\prompt{[QED]0.78[/QED]}. \\

\noindent This methodology is inspired by the calibration analysis in the GPT-4 technical report \citep{gpt4}, which highlights calibration as a key indicator of high-quality pretraining. For each response, we calculated the model's predicted probability from the perplexity of the text, normalizing it against other responses for the same question. These probabilities were then aggregated and sorted into 10 equal-width bins. We plotted the fraction of correct responses for each bin, allowing us to visualize the relationship between the model's confidence and accuracy.

\subsubsection{Results}
\begin{figure}[t]
    \centering
    \begin{subfigure}[b]{0.46\textwidth}
        \centering
        \includegraphics[width=\textwidth]{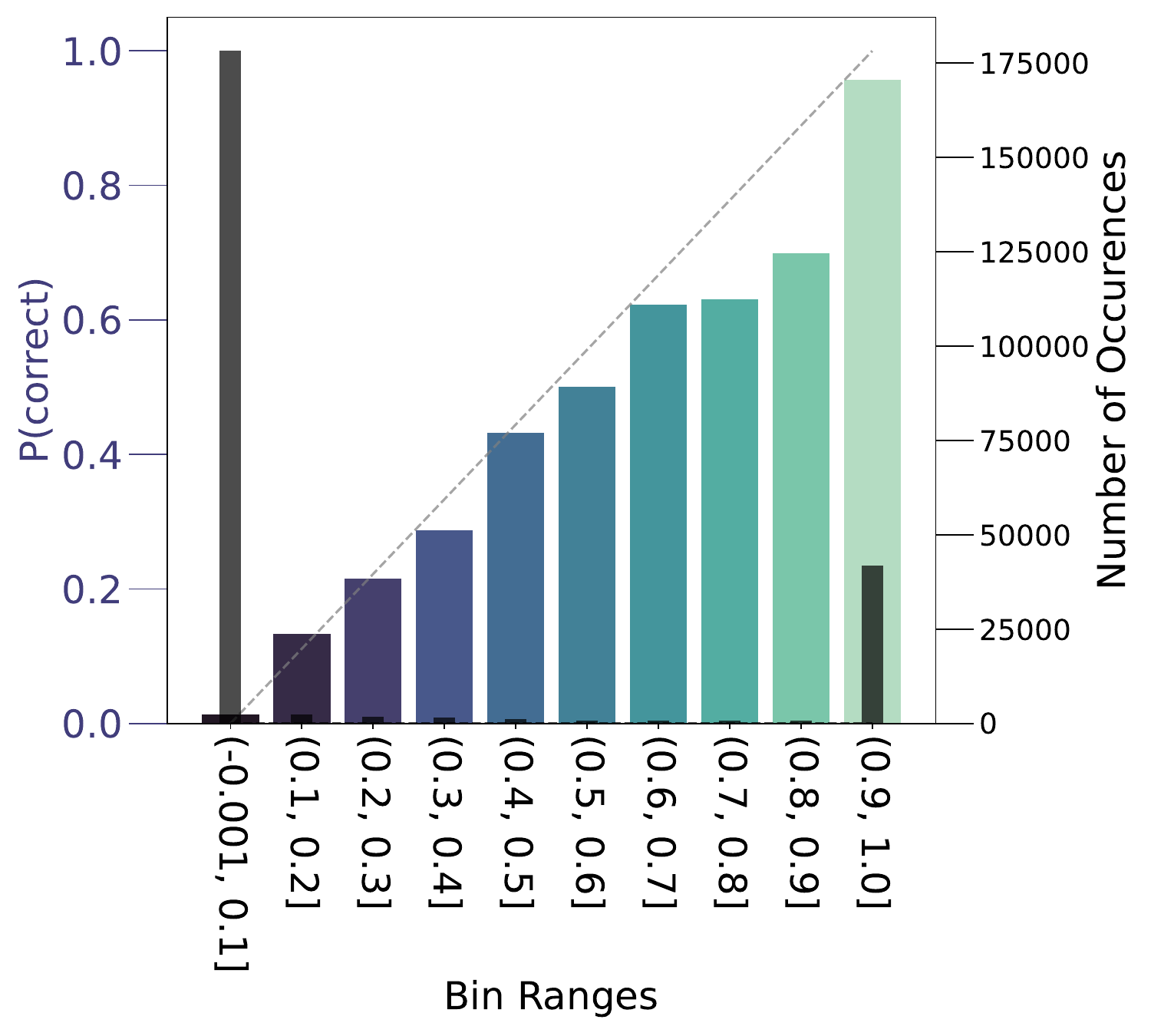}
        \caption{Calibration of Chemma-2B.}
        \label{fig:calibration-2b}
    \end{subfigure}
    \hfill
    \begin{subfigure}[b]{0.46\textwidth}
        \centering
        \includegraphics[width=\textwidth]{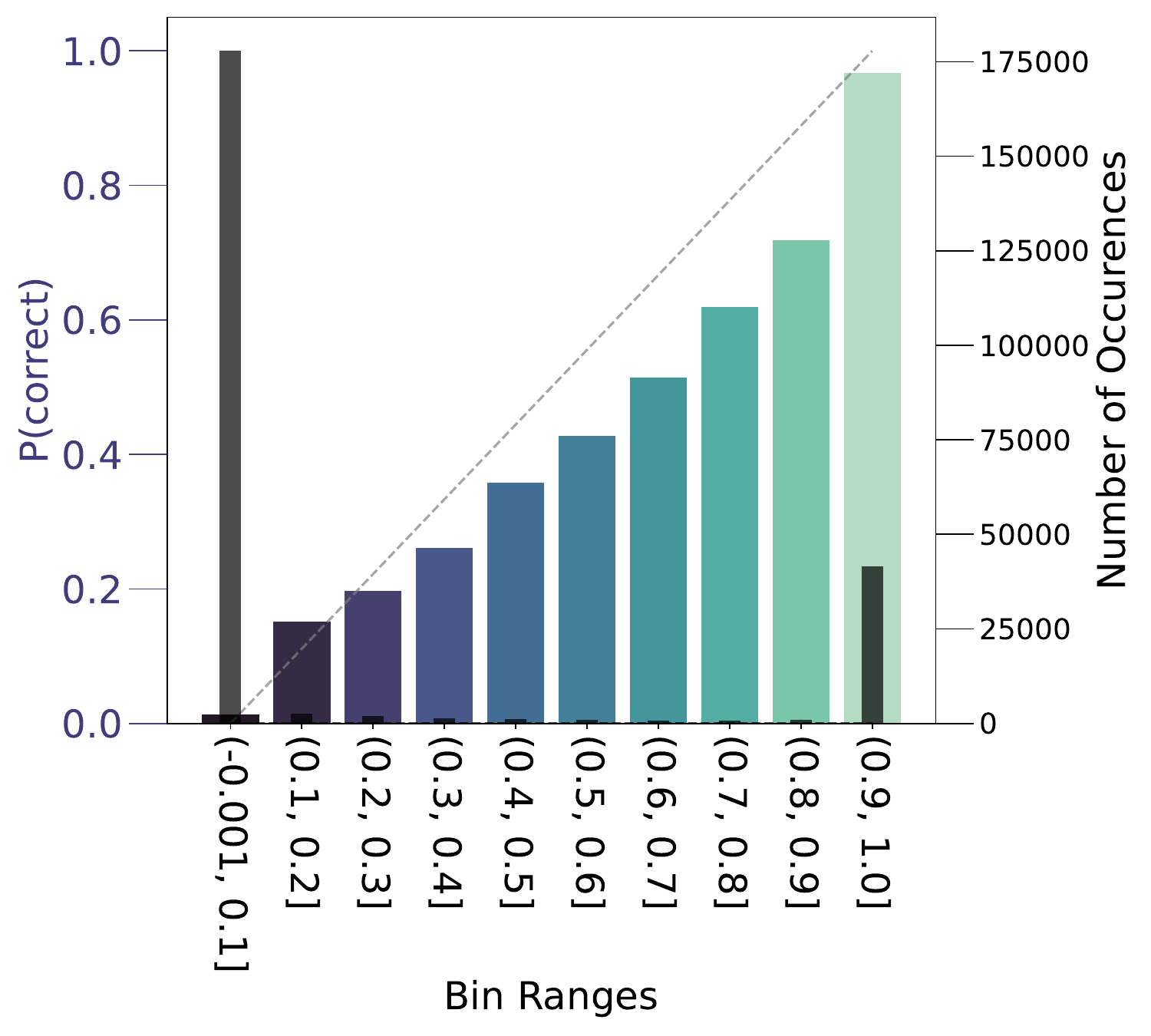}
        \caption{Calibration of Chemlactica-125M.}
        \label{fig:calibration-125m}
    \end{subfigure}
    \caption{Model calibration on synthetic multiple choice question where y=x represents perfect calibration.}
    \label{fig:calibrations}
\end{figure}

Figures \ref{fig:calibration-2b} and \ref{fig:calibration-125m} present the calibration plots for Chemma-2B and Chemlactica-125M, respectively. The x-axis represents the 10 probability bins, while the left y-axis shows the correct response fraction. The right y-axis and red bars indicate the number of occurrences within each bin.\\

\noindent Chemlactica and Chemma models demonstrate robust calibration, as evidenced by the near-linear relationship between assigned probabilities and correct outcomes across all computed properties. This relationship closely follows the diagonal grey line, which represents perfect calibration.\\

\noindent These results suggest that the perplexity scores generated by our models serve as reliable confidence indicators for molecular data predictions (averaged over a set of molecules), provided the data falls within the distribution of the training corpus. This calibration is crucial for practical applications, as it allows users to accurately gauge the reliability of the models' outputs in simple molecular prediction and generation tasks. However, finetuning, like that performed in the optimization algorithm, likely leads to a loss of model calibration\citep{gpt4}.

\subsection{Additional Figures}
\subsubsection{Visualization of the Model Outputs on Property Prediction and Conditional Generation Tasks}

\begin{figure}
    \centering
    \begin{subfigure}[b]{0.46\textwidth}
        \centering
        \includegraphics[width=\textwidth]{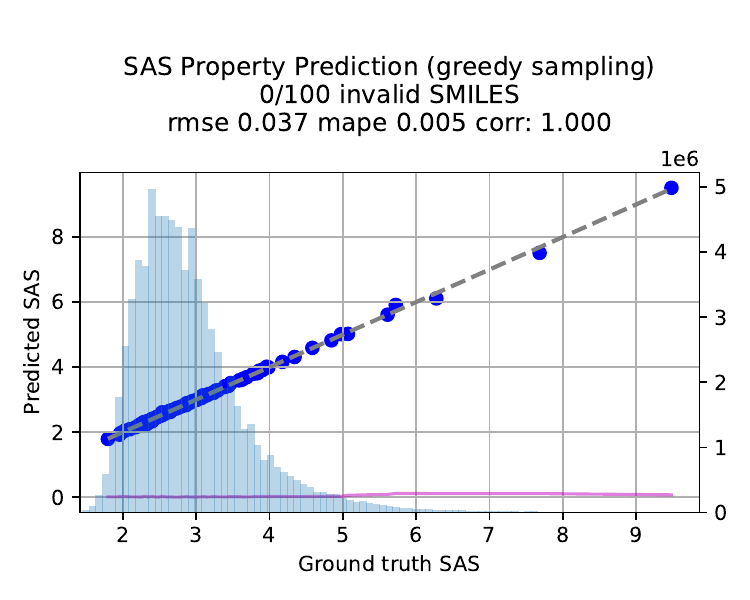}
        \caption{SAS prediction.}
        \label{fig:PP SAS-2b}
    \end{subfigure}
    \hfill
    \begin{subfigure}[b]{0.46\textwidth}
        \centering
        \includegraphics[width=\textwidth]{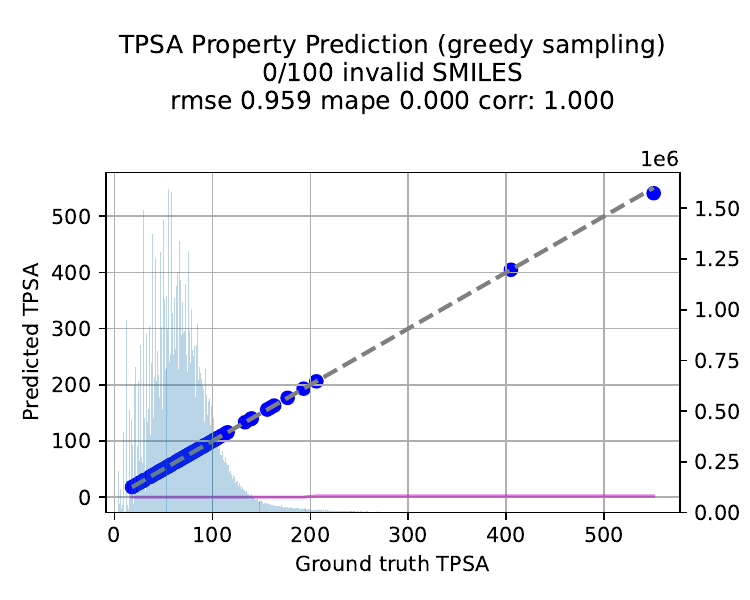}
        \caption{TPSA Prediction.}
        \label{fig:PP TPSA-2b}
    \end{subfigure}

    \vspace{0.25em}

    \begin{subfigure}[b]{0.46\textwidth}
        \centering
        \includegraphics[width=\textwidth]{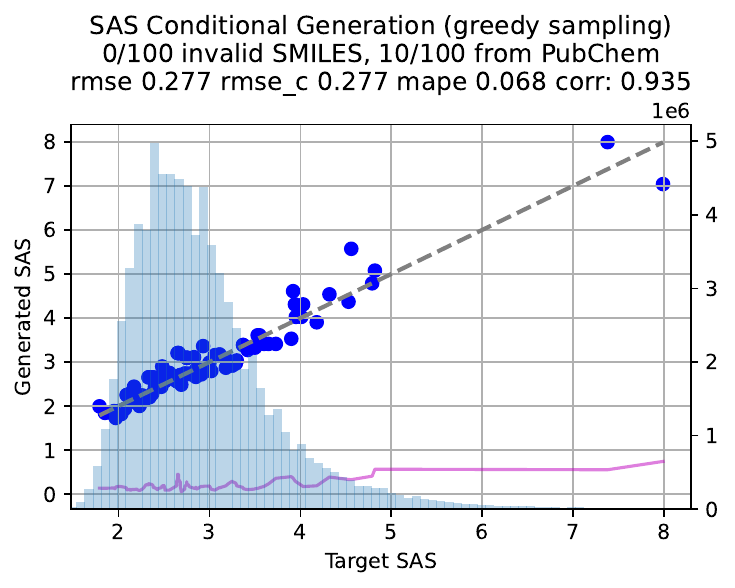}
        \caption{SAS-conditioned generation of molecules.}
        \label{fig:CG SAS-2b}
    \end{subfigure}
    \hfill
    \begin{subfigure}[b]{0.46\textwidth}
        \centering
        \includegraphics[width=\textwidth]{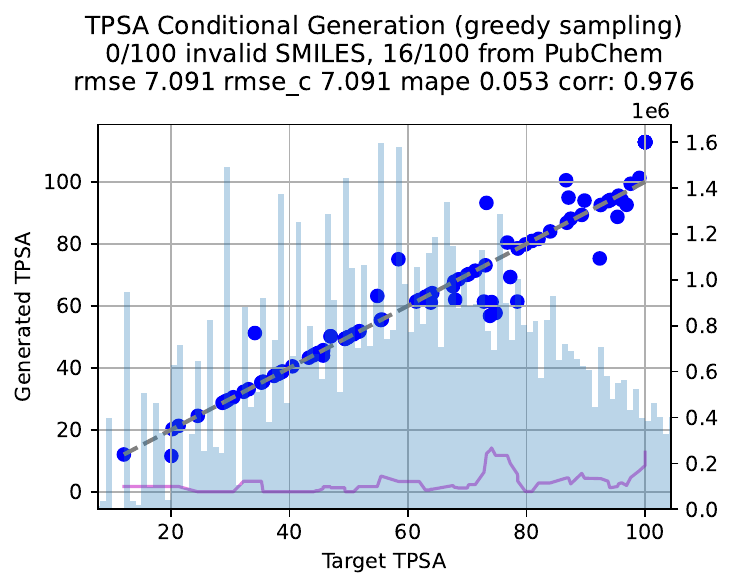}
        \caption{TPSA-conditioned generation of molecules.}
        \label{fig:CG TPSA-2b}
    \end{subfigure}
    
    \vspace{0.25em}
    
    \begin{subfigure}[b]{0.46\textwidth}
        \centering
        \includegraphics[width=\textwidth]{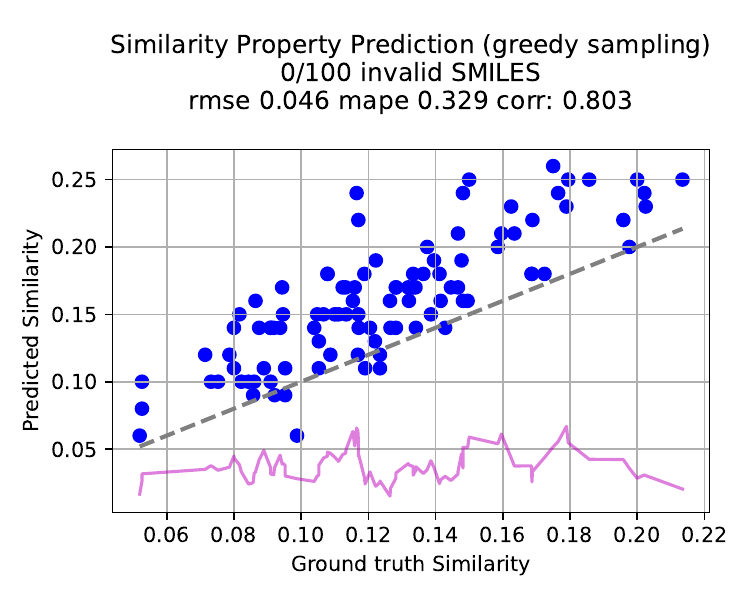}
        \caption{Prediction of similarity between two molecules.}
        \label{fig:PP SIM-2b}
    \end{subfigure}
    \hfill
    \begin{subfigure}[b]{0.46\textwidth}
        \centering
        \includegraphics[width=\textwidth]{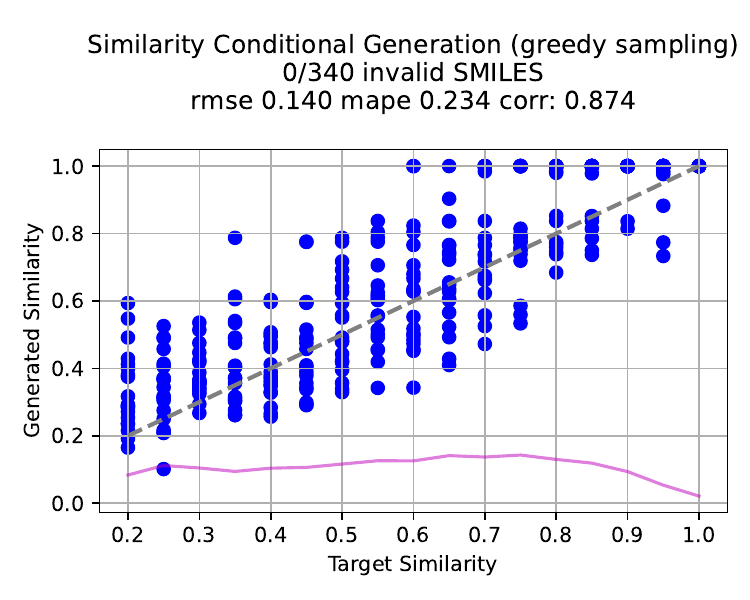}
        \caption{Similarity-conditioned generation of molecules.}
        \label{fig:CG SIM-2b}
    \end{subfigure}
    \caption{Illustration of errors made by Chemma-2B during property prediction and conditional generation for various properties.}
    \label{fig:chemma-2b-results}
\end{figure}

Figures \ref{fig:PP SIM-2b}-\ref{fig:PP SIM-2b} show the performance of Chemma-2B for property prediction and conditional molecular generation tasks. Each dot in the scatter plot corresponds to one molecule. The histogram in the background is the distribution of those properties in our training set. The purple line represents the RMSE between the property's ground truth and predicted values.


\FloatBarrier
\clearpage
\FloatBarrier
\subsubsection{Docking Scores Throughout DRD2 MPO}
\label{sec:dockscoresonly}
\begin{figure}[!ht]
    \centering
    \includegraphics[width=0.95\linewidth]{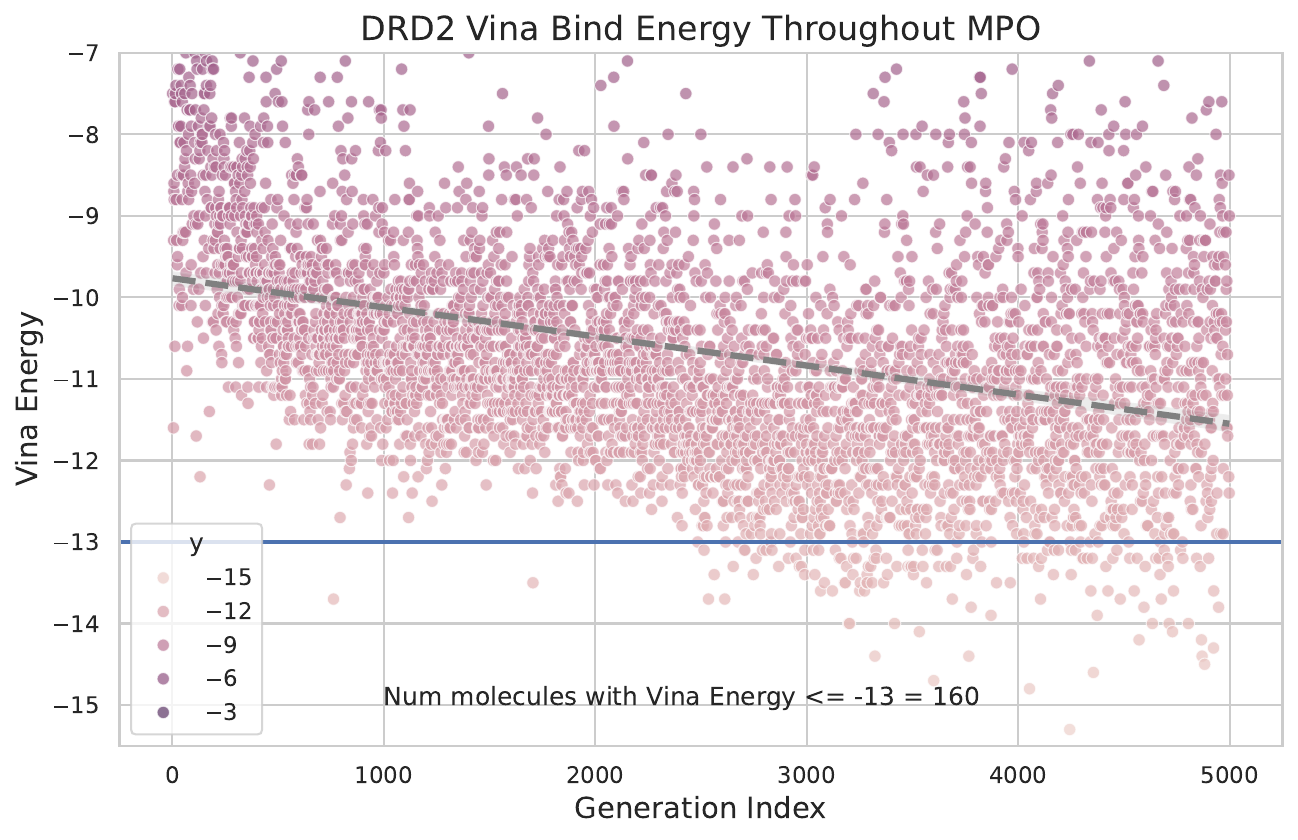}
    \caption{Autodock Vina energy scores for molecules generated through DRD2 MPO process; note that molecules which did not dock at all are not included.}
\end{figure}

\FloatBarrier
\subsubsection{Generated Molecules from the Docking Experiments}
\label{sec:docking-molecules-examples}

\begin{figure}[!ht]
    \centering
    \includegraphics[width=0.95\linewidth]{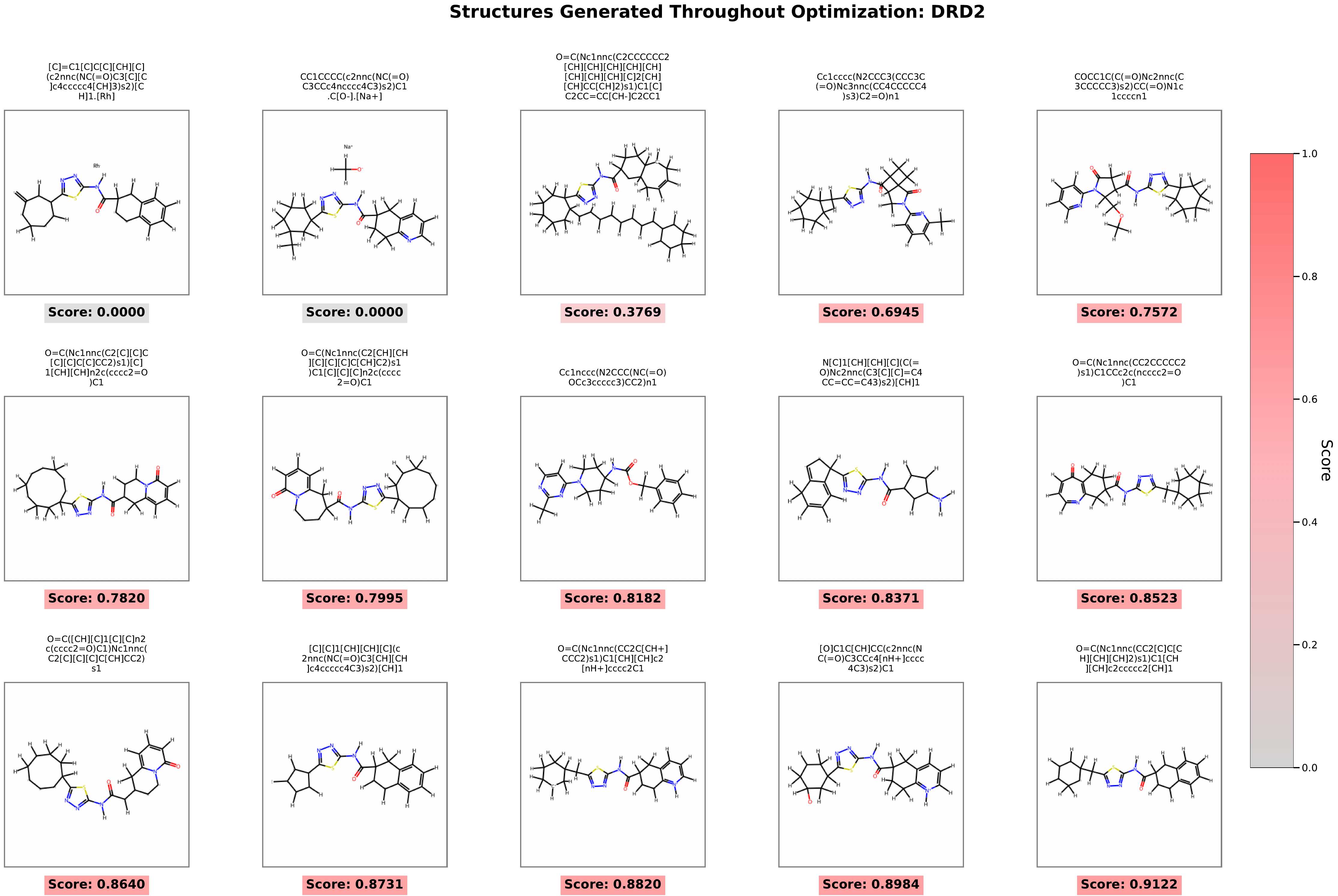}
\end{figure}
\vfill
\clearpage

\begin{figure}[!ht]
    \centering
    \includegraphics[width=1.0\linewidth]{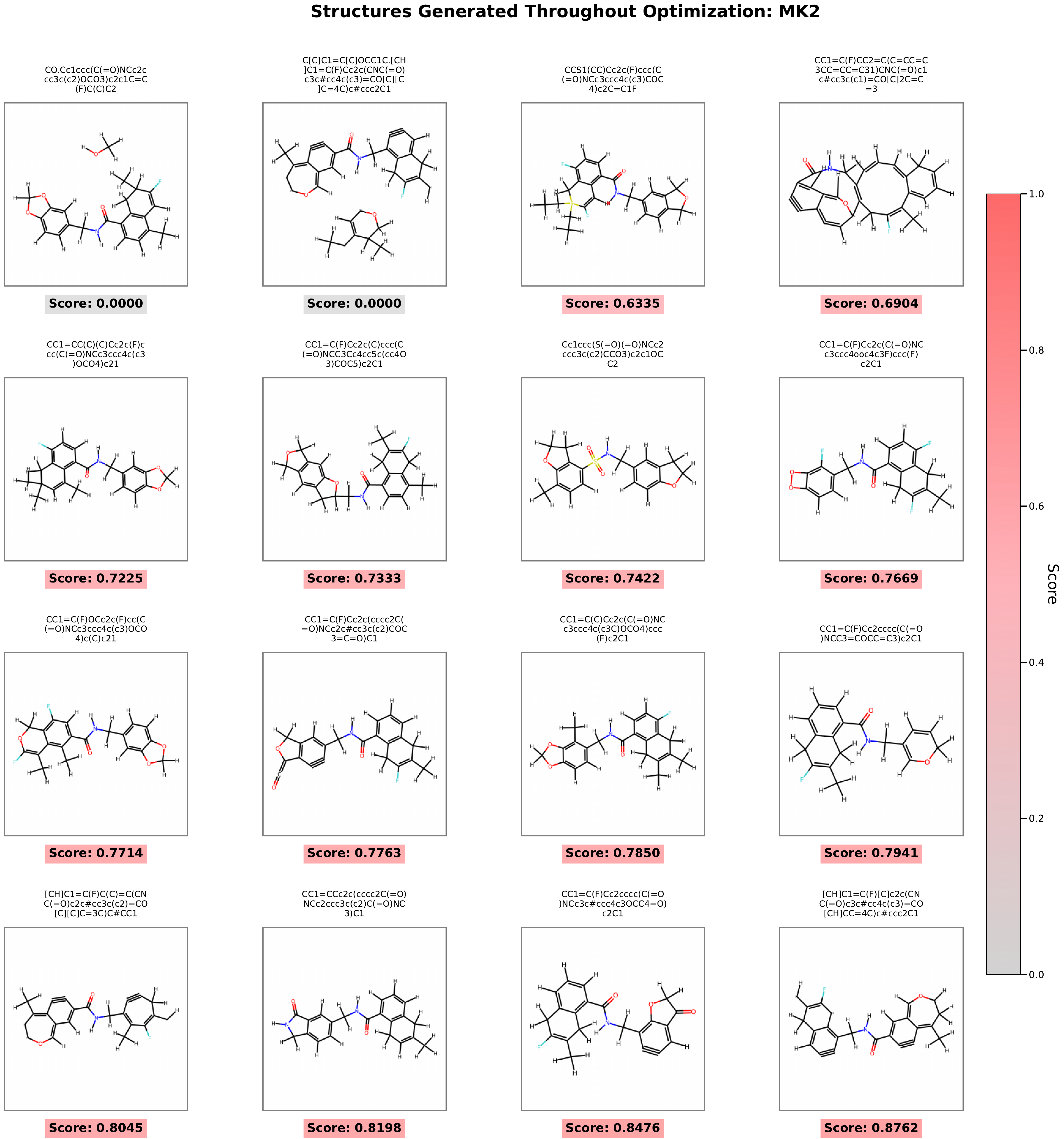}
\end{figure}
\vfill
\clearpage

\begin{figure}[!ht]
    \centering
    \includegraphics[width=1.0\linewidth]{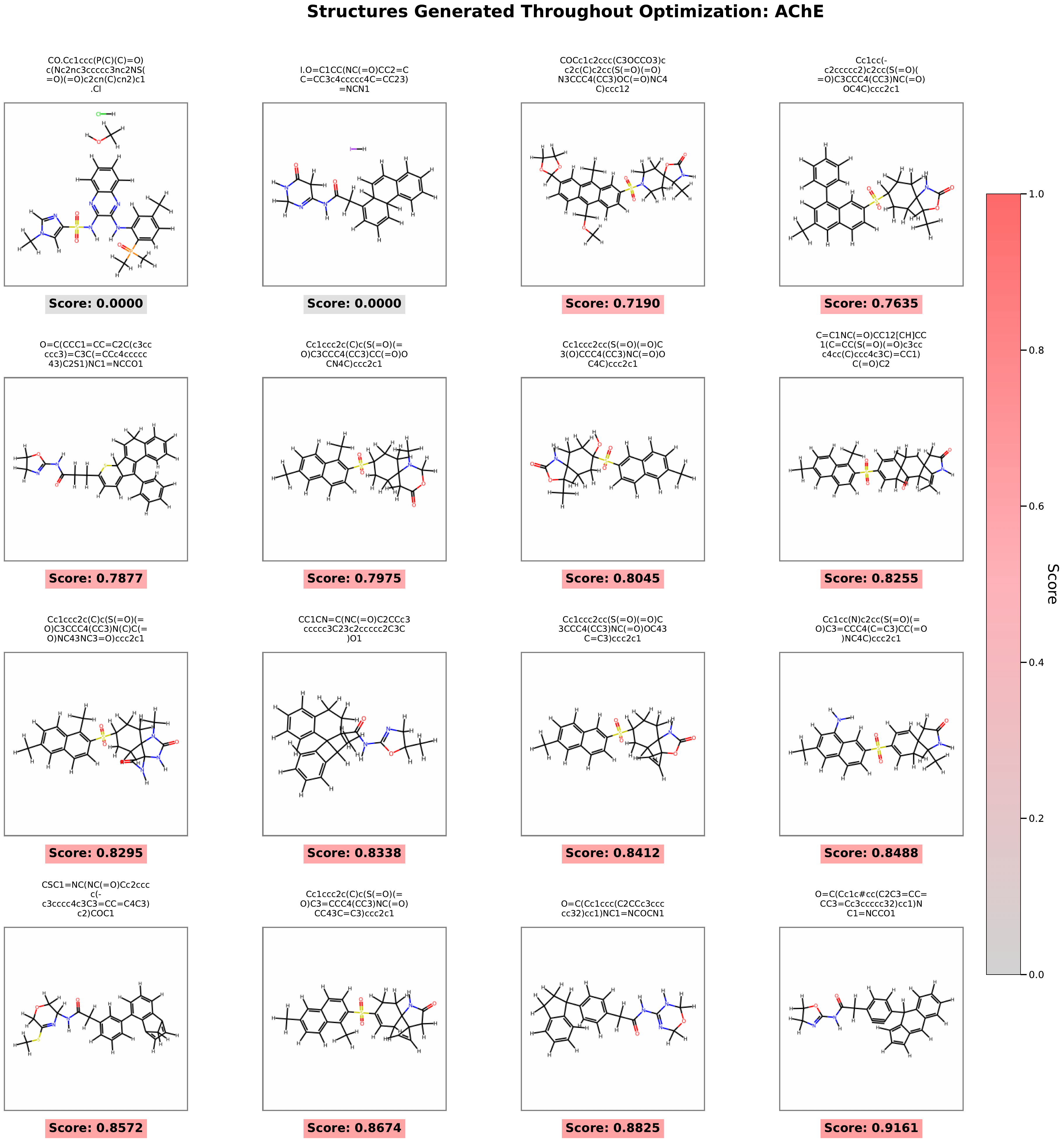}
\end{figure}
\vfill
\clearpage





\bibliography{chem}

\end{document}